 \documentclass[sigconf]{acmart}
\AtBeginDocument{%
  }

\setcopyright{acmlicensed}
\copyrightyear{2018}
\acmYear{2018}
\acmDOI{XXXXXXX.XXXXXXX}
\acmConference[Conference acronym 'XX]{Make sure to enter the correct
  conference title from your rights confirmation email}{June 03--05,
  2018}{Woodstock, NY}
\acmISBN{978-1-4503-XXXX-X/2018/06}

\usepackage{amsmath} 
\usepackage{amsthm}
\usepackage{enumitem}
\usepackage{multirow}
\usepackage{booktabs}
\usepackage{makecell}  
\usepackage{array} 
\usepackage{tabularx}
\usepackage[table]{xcolor}
\usepackage{colortbl}
\usepackage{balance}

\begin{document}
\copyrightyear{2026}
\acmYear{2026}
\setcopyright{cc}
\setcctype{by}
\acmConference[KDD '26]{Proceedings of the 32nd ACM SIGKDD Conference on Knowledge Discovery and Data Mining V.2}{August 09--13, 2026}{Jeju Island, Republic of Korea}
\acmBooktitle{Proceedings of the 32nd ACM SIGKDD Conference on Knowledge Discovery and Data Mining V.2 (KDD '26), August 09--13, 2026, Jeju Island, Republic of Korea}
\acmDOI{10.1145/3770855.3818084}
\acmISBN{979-8-4007-2259-2/2026/08}

\title{Outsmarting the Chameleon: Counterfactual Decoupling for Tactical OOD Shifts in Live Streaming Risk Assessment}

\author{Yiran Qiao}
\authornote{This work was conducted during Yiran's internship at ByteDance China.}
\authornotemark[3]
\authornotemark[4]
\orcid{0009-0006-0632-0066}
\affiliation{%
  \institution{Institute of Computing Technology, Chinese Academy of Sciences}
  \city{Beijing}
  \country{China}
}
\email{yrqiao@gmail.com}

\author{Jing Chen}
\orcid{0000-0003-2672-4587}
\affiliation{%
  \institution{ByteDance China}
  \city{Hangzhou}
  \country{China}
}
\email{yilan.chan@bytedance.com}

 \author{Jiaqi Xu}
\authornotemark[3]
\authornotemark[4]
\orcid{0009-0000-4170-2648}
\affiliation{%
  \institution{Institute of Computing Technology, Chinese Academy of Sciences}
  \city{Beijing}
  \country{China}
}
\email{xujiaqi253@mails.ucas.ac.cn}

 \author{Yang Liu}
 \authornotemark[3]
\authornotemark[4]
 \orcid{0000-0002-1525-0788}
\affiliation{%
  \institution{Institute of Computing Technology, Chinese Academy of Sciences}
  \city{Beijing}
  \country{China}
}
\email{liuyang2023@ict.ac.cn}

\author{Qiwei Zhong}
\orcid{0000-0002-8517-8072}
\affiliation{%
  \institution{ByteDance China}
  \city{Hangzhou}
  \country{China}
}
\email{huafeng.hf@bytedance.com}

\author{Xiang Ao}
\authornote{Corresponding author.}
\authornote{State Key Laboratory of AI Safety, Institute of Computing Technology, Chinese Academy of Sciences.}
\authornote{Also with University of Chinese Academy of Sciences.}
\orcid{0000-0001-9633-8361}
\affiliation{%
  \institution{Institute of Computing Technology, Chinese Academy of Sciences}
  \city{Beijing}
  \country{China}
}
\email{aoxiang@ict.ac.cn}

\renewcommand{\shortauthors}{Yiran Qiao et al.}

\begin{abstract}
Live streaming has emerged as a primary medium for social interaction and digital commerce, yet it is increasingly plagued by sophisticated risks. A fundamental challenge in this domain is \emph{tactical out-of-distribution (OOD) shift}: while malicious actors maintain stable underlying objectives, they continuously redesign narrative packaging to evade detection. 
Such adversarial shifts expose critical limitations of existing OOD generalization paradigms, whose assumptions are difficult to satisfy in the presence of tightly coupled intent–tactic evolution and ill-defined raw-level counterfactuals.

In this paper, we tackle this issue from a \emph{latent causal} perspective and propose \underline{L}atent-\underline{P}redictive \underline{C}ounterfactual \underline{D}ecoupling~(LPCD), a plug-in framework for robust live streaming risk assessment. LPCD enables counterfactual reasoning under adversarial tactical re-packaging by modeling intent and narrative variation at the latent level, and enforces \emph{latent counterfactual consistency} to anchor risk prediction on causally stable malicious intent. At inference time, LPCD applies a lightweight, parameter-free calibration to further mitigate tactic-induced distribution shifts. Extensive experiments on large-scale industrial datasets and online production traffic demonstrate that LPCD consistently outperforms state-of-the-art baselines, validating its effectiveness in moderating evolving adversarial risks in real-world live streaming. The project page is available at \url{https://qiaoyran.github.io/LiveStreamingRiskAssessment/}.

\end{abstract}



\maketitle

\newcommand\kddavailabilityurl{https://doi.org/10.5281/zenodo.20446272}
\ifdefempty{\kddavailabilityurl}{}{
\begingroup\small\noindent\raggedright\textbf{Resource Availability:}\\
The source code of this paper has been made publicly available at \url{https://doi.org/10.5281/zenodo.20446272}.
\endgroup
}

\section{Introduction}

Live streaming has become a primary medium for social interaction and digital commerce, accompanied by increasingly sophisticated risks such as financial fraud and illicit promotion. Malicious behaviors in these sessions are often embedded within socially plausible narratives, which conceal true objectives and make detection challenging. These diverse surface behaviors often mask a small set of stable malicious objectives, allowing adversaries to adapt their tactics over time without altering the underlying intent.

\begin{figure}[b]
\centering
\includegraphics[width=0.45\textwidth]{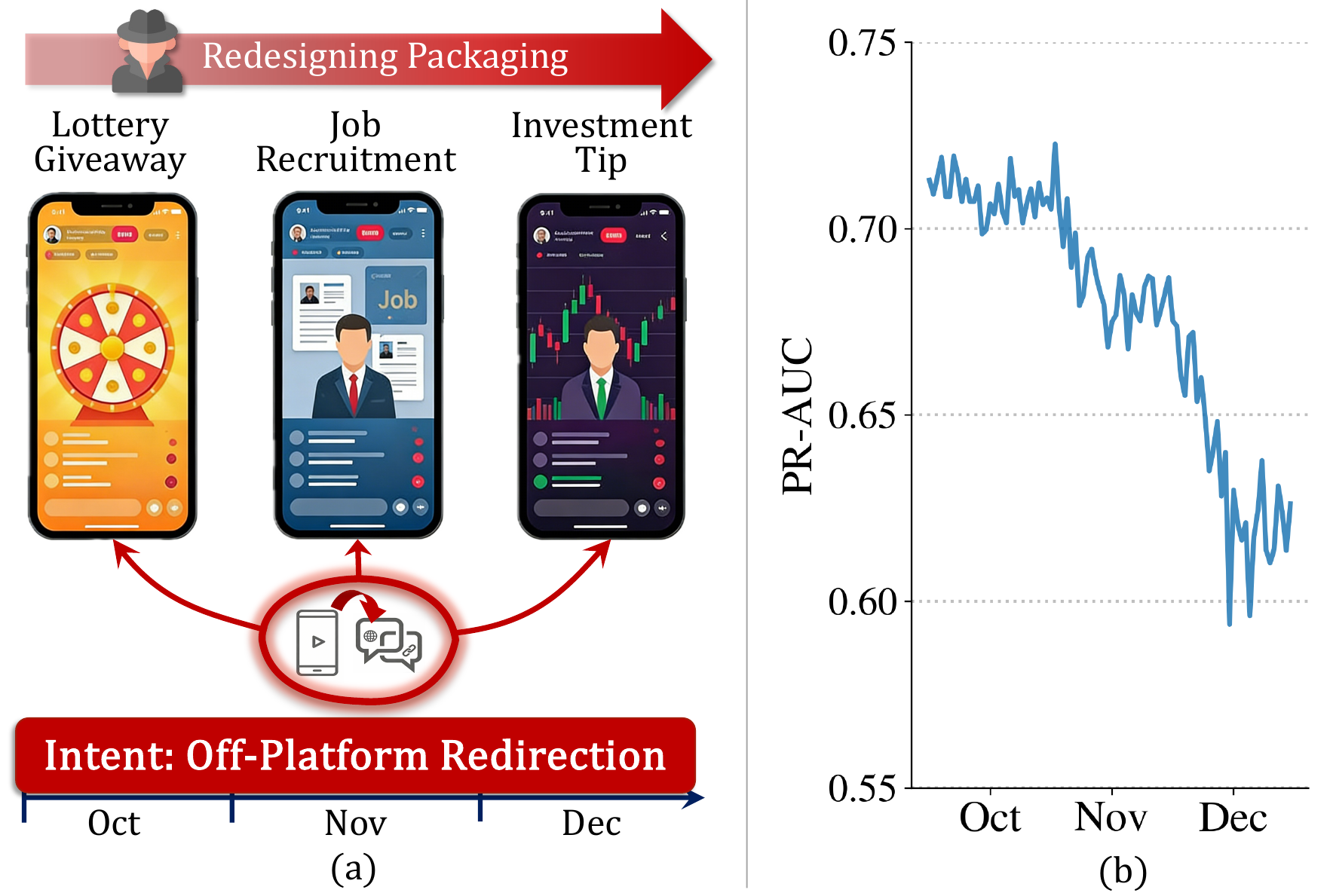}
\caption{(a) Adversaries maintain an invariant malicious intent (e.g., off-platform redirection) while continuously redesigning the volatile narrative packaging to evade detection.
(b) PR-AUC of a production risk detection model evaluated on real-world data from October to December 2025, showing a degradation in performance over the same period.}
\label{fig:intro}
\end{figure}

A dominant class of objectives includes \textit{(i) off-platform redirection} to external scam environments and \textit{(ii) on-platform deceptive monetization} through fraudulent sales. To achieve these objectives under scrutiny, adversaries continuously redesign the narrative packaging of a live session, including conversational scripts, interaction rhythms, and coordination between hosts and accomplices. For instance, the same redirection intent may be framed as a lottery giveaway, a job recruitment, or an investment tip, as illustrated in Figure~\ref{fig:intro}(a). The resulting mismatch between stable intent and volatile presentation creates a persistent challenge for models that attempt to generalize from historical patterns.

This phenomenon constitutes a \textbf{tactical out-of-distribution (OOD) shift}, where the data distribution changes at a strategic level while the underlying risk-generating logic remains invariant.
Unlike conventional distribution shifts driven by passive or exogenous factors, tactical OOD shifts arise from adversarially optimized narrative redesigns that are intentionally coupled with the malicious objective. Consequently, models that rely on historical tactical patterns often fail to generalize when a known intent is wrapped in an unseen narrative, as shown in Figure~\ref{fig:intro}(b). 

Despite extensive research on OOD generalization~\cite{zhou2022domain,liu2021towards}, existing approaches face fundamental limitations when applied to live streaming risk assessment. At the supervision level, most OOD methods rely on explicit~\cite{arjovsky2019invariant, krueger2021out} or implicitly inferable environment labels~\cite{creager2021environment,liu2024time}. In live streaming, however, tactical variations emerge continuously and adversarially, without clear environment boundaries. This makes it difficult to directly apply environment-based invariance assumptions in practice.

Beyond this supervision challenge, adversarial live streaming violates a key assumption shared by many invariance-based methods. These approaches typically presume that spurious correlations arise from passive or weakly coupled shifts~\cite{zhang2020causal, liu2021learning}. In contrast, narrative packaging in malicious live sessions is strategically designed and tightly coupled with underlying intent. This strategic co-evolution leads to deep semantic entanglement, under which enforcing invariance at the observation level can be insufficient and, in some cases, even counterproductive. 


While counterfactual reasoning~\cite{pearl2009causality,feder2022causal} offers a principled path to address such entanglement, constructing realistic counterfactuals within the raw observation space is often ill-defined in practice. Live sessions comprise high-dimensional, multimodal streams, where input-level interventions are difficult to specify without violating semantic coherence. These challenges motivate a latent causal formulation, in which counterfactual reasoning and invariance are enforced in the latent representation space rather than on raw observations.

To this end, we advocate a \emph{latent causal} perspective that enables counterfactual reasoning under adversarial tactical re-packaging.
As raw-level counterfactuals are ill-defined for live sessions, we perform causal interventions in the latent representation space, where intent-preserving tactical variations can be explicitly modeled.
This structure allows us to enforce latent counterfactual consistency, ensuring the model remains focused on the invariant risk core despite strategic narrative changes.

Building on this perspective, we propose \textbf{Latent-Predictive Counterfactual Decoupling (LPCD)}, a plug-in framework for robust live streaming risk assessment.
LPCD models session representations as composed of intent-related and packaging-related factors, and enforces \emph{latent counterfactual consistency} by intervening on the packaging factor during training, thereby isolating intent-specific signals that remain causally stable under tactical re-packaging.
At test time, LPCD further applies a parameter-free calibration to rectify tactic-induced magnitude shifts.
Extensive experiments on large-scale industrial data from Douyin show that LPCD consistently outperforms strong baselines in both in-distribution and tactical OOD settings.
Our main contributions are summarized as follows:
\begin{itemize}[leftmargin=*,topsep=5pt]

\item We identify \emph{tactical out-of-distribution (OOD) shift} as a fundamental challenge in live streaming risk assessment, characterized by invariant malicious intent under adversarially evolving narrative packaging, and provide a principled framing from a \emph{latent causal} perspective.

\item We propose \textbf{Latent-Predictive Counterfactual Decoupling (LPCD)}, a plug-in framework that enforces latent counterfactual consistency by intervening on narrative packaging at both the representation and prediction levels, enabling intent-focused risk modeling. 

\item Extensive experiments on large-scale industrial live-streaming datasets and online validation confirm LPCD's SOTA performance in both in-distribution and tactical OOD settings, validating its efficacy in moderating evolving adversarial risks in real-world live streaming.
\end{itemize}

\section{Related Work}
\subsection{Risk Assessment in Online Ecosystems}
Risk assessment in online ecosystems has evolved from fine-grained artifact detection to more holistic modeling of coordinated behaviors. One line of research focuses on identifying localized signals, such as toxic language in user-generated text~\cite{lees2022new,zannettou2020measuring} or policy-violating visual cues in short videos~\cite{lu2025vlm,wang2025reasoning}. To capture more complex and organized risks, another line adopts sequential~\cite{guo2018learning,qiao2025online,xiao2024vecaug,qiao2024Financial,wang2023sequence} and graph-based models~\cite{dou2020enhancing,huang2022auc,shi2022h2,li2021live,cheng2025graph}, enabling the characterization of temporal dependencies and cross-entity coordination.

In live streaming, risk signals are inherently session-level, emerging from long-range interactions and evolving narratives rather than isolated events. This has led to Multiple Instance Learning (MIL) formulations, exemplified by AC-MIL~\cite{qiao2026livelieactionawarecapsule}, which models live sessions as collections of user–timeslot instances under session-level supervision. While such approaches effectively capture intra-session dynamics, they remain largely associative, entangling risk predictions with surface narrative patterns.

Under adversarial tactic evolution, where identical malicious intents are repeatedly rewrapped in novel narratives, this coupling therefore limits robustness to tactical distribution shifts, motivating the need for intent-focused modeling beyond holistic session representations.

\subsection{Causal Perspectives on OOD Generalization}
Prior work on out-of-distribution (OOD) generalization aims to improve robustness by enforcing invariant representations across environments~\cite{arjovsky2019invariant, krueger2021out, Sagawa*2020Distributionally, zhou2022domain, liu2021towards}.
Causality-inspired approaches further interpret distribution shifts as interventions on non-causal factors, and seek to disentangle causal semantics from spurious correlations~\cite{zhang2020causal, liu2021learning, mahajan2021domain}.

However, most existing frameworks operate under a passive or exogenous shift assumption, where variations arise from low-level statistical noise, backgrounds, or temporal non-stationarity~\cite{oublal2024disentangling, liu2025long, wu2025out}.
In these scenarios, task semantics are typically assumed to be stable and counterfactual variations are treated as well-defined at the observation level, with distribution shifts viewed as environment-induced rather than strategic.

In contrast, live streaming risk assessment operates in a tactical OOD regime.
Malicious actors actively redesign narrative packaging, interaction patterns, and temporal strategies to obscure intent.
These shifts are structured, high-dimensional, and intentionally entangled with risk signals, going beyond the scope of prior methods that focus on attribute-level disentanglement or statistical invariance.
Our work addresses this gap by introducing a latent counterfactual decoupling framework that explicitly intervenes on narrative packaging, enabling robust intent inference under evolving adversarial tactics.

\section{Problem Formulation}

\subsection{Business Setting}
Live streaming platforms face \emph{adversarially evolving risks} where malicious actors continuously re-engineer tactics to evade detection. This environment presents three critical challenges: (1)~\textbf{Tactical shifts}: Surface-level narrative packaging and interaction scripts evolve rapidly, while the underlying malicious intent remains invariant. (2)~\textbf{Coarse supervision}: Only session-level labels are available without explicit environment or action-level annotations, complicating group-aware OOD schemes. (3)~\textbf{Label latency}: Delays in manual reviews create a temporal gap between live events and label availability, necessitating models that generalize across distribution shifts without real-time retraining.

\subsection{Definition and Objective}

We study the \emph{live streaming risk assessment} problem under tactical OOD shifts. The goal is to determine whether a live streaming session involves risky behaviors such as fraud or illicit promotion, despite evolving tactics designed to evade detection.

\begin{definition}
\textbf{(Action)}
An \emph{action} in a live streaming session is represented as a tuple
$
\alpha = (u, t, a, x),
$
where $u$ denotes the user performing the action, $t$ is the timestamp, $a$ indicates the action type (e.g., message posting, gifting, joining), and $x \in \mathbb{R}^d$ is a $d$-dimensional semantic embedding extracted from the raw textual content using a pretrained language model. 
\end{definition}

\begin{definition}
\textbf{(Live Streaming Session)}  
A live streaming session over a time window $[0,T]$ is defined as
\[
S^{[0,T]} = \big(\mathcal{U}, [\alpha_1, \alpha_2, \ldots, \alpha_N]\big),
\]
where $\mathcal{U} = \{u^{\mathrm{h}}\} \cup U^{\mathrm{v}}$ consists of a unique host $u^{\mathrm{h}}$ and a set of participating viewers, and
$[\alpha_1, \alpha_2, \ldots, \alpha_N]$ is the chronologically ordered sequence of actions within $[0,T]$.
 Each action $\alpha_i$ implicitly carries user and temporal context through $(u_i, t_i)$.
\end{definition}

\begin{definition}
\textbf{(Live Streaming Session Encoder)} 
\label{def:encoder}
In practice, risk assessment models typically rely on an intermediate session-level representation that aggregates information across all actions.
We therefore assume a generic backbone encoder
\[
\mathcal{E}(\cdot): S^{[0,T]} \rightarrow \mathbf{x} \in \mathbb{R}^D,
\]
which maps a live streaming session to a $D$-dimensional embedding $\mathbf{x}$.
The encoder $\mathcal{E}(\cdot)$ can be instantiated by any sequence or multi-instance learning~(MIL) model, and is trained jointly with the downstream risk predictor.
Our method operates as a plug-in module on top of this session representation, without imposing architectural constraints on $\mathcal{E}(\cdot)$.
\end{definition}
\noindent
\textbf{Problem Objective.}
Given a dataset
$\mathcal{D} = \{(S_i^{[0,T]}, y_i)\}_{i=1}^N,
$
where $y_i \in \{0,1\}$ indicates whether session $i$ is risky, the objective is to learn a function
$f: S^{[0,T]} \rightarrow [0,1],
$
that estimates the probability that a session involves malicious activity.

\section{Methodology}
\begin{figure*}[htbp]
  \centering
  \includegraphics[width=\textwidth]{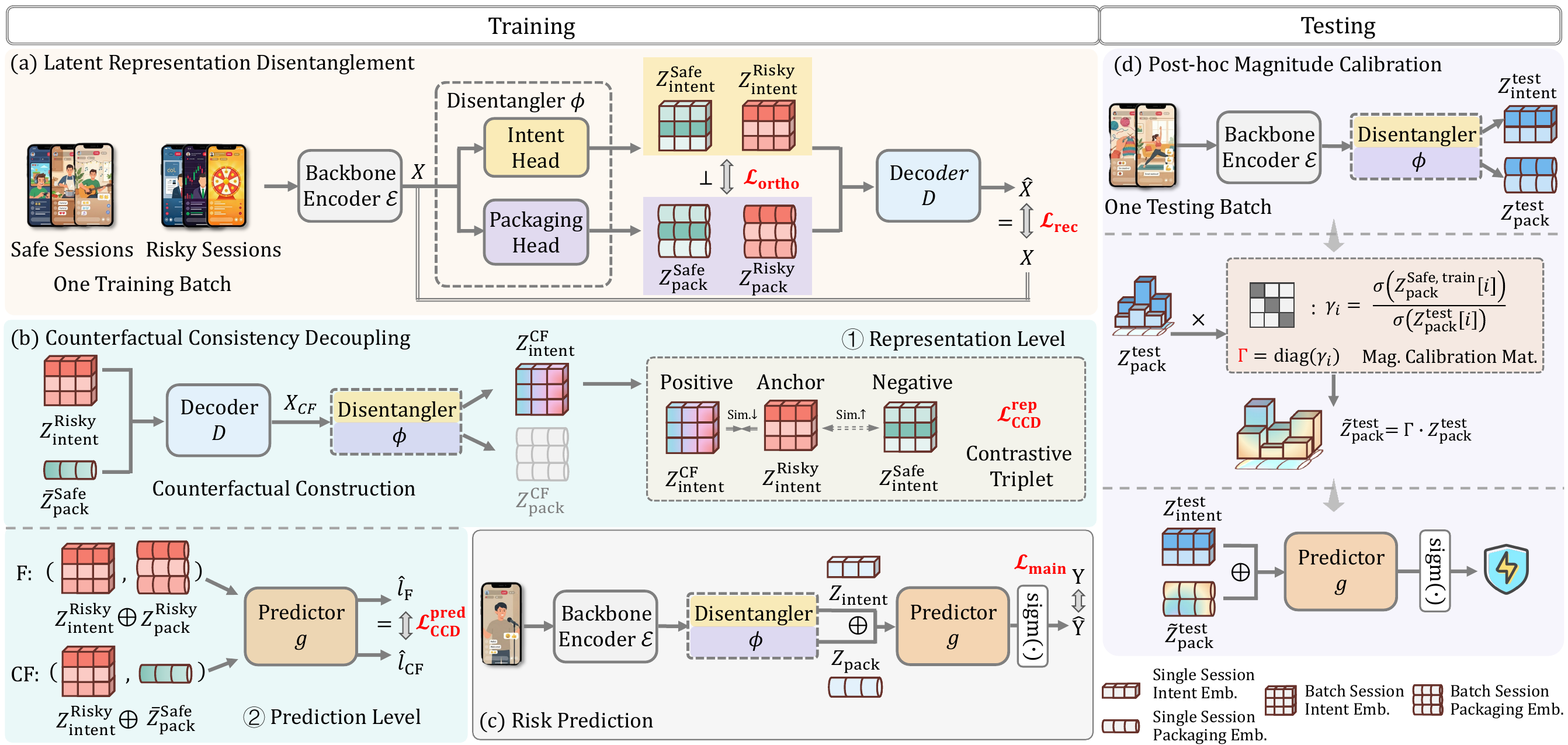}

  \caption{ Overview of LPCD.
In training flow:
(a) Latent Representation Disentanglement factorizes session representations into intent and packaging components;
(b) Counterfactual Consistency Decoupling enforces intent invariance under counterfactual packaging at both the representation and prediction levels;
and (c) Risk Prediction aggregates the disentangled factors to produce session-level risk scores.
At test time, (d) Post-hoc Magnitude Calibration adjusts tactic-induced magnitude shifts in packaging representations before inference, enabling robust deployment under evolving adversarial tactics.
}
  \label{fig:overview}
\end{figure*}

\subsection{Overview of LPCD}
Figure~\ref{fig:overview} presents an overview of our proposed LPCD framework for live streaming risk assessment, which combines latent causal decoupling with post-hoc magnitude calibration.

As illustrated in Figure~\ref{fig:overview}, this plug-in framework consists of three training-stage components and a lightweight inference-stage calibration module.
In training flow: \textbf{(a) Latent Representation Disentanglement} decomposes session representations into intent and packaging factors, capturing underlying malicious objectives and their tactical realizations, respectively.
\textbf{(b) Counterfactual Consistency Decoupling} enforces intent invariance under counterfactual packaging at both the representation and prediction levels, mitigating spurious correlations induced by tactic evolution.
\textbf{(c) Risk Prediction} aggregates the disentangled factors to produce session-level risk scores under standard supervision.
At test time, \textbf{(d) Post-hoc Magnitude Calibration} further rectifies tactic-induced magnitude shifts in packaging representations at test time before risk inference.

This design enables robust risk prediction by isolating stable malicious intent, decoupling tactical variations, and correcting distributional drift during deployment.

\subsection{Latent Representation Disentanglement}

Existing live streaming risk assessment models~\cite{qiao2026livelieactionawarecapsule} represent each session using a single embedding, which naturally entangles malicious intent with tactical packaging.
Under evolving narrative strategies, such entanglement complicates identifying intent-relevant signals that remain stable across tactical variations.
To expose these stable factors and enable latent counterfactual analysis, we decompose the session representation into intent-related and packaging-related factors, following the principles of disentangled representation learning works~\cite{higgins2017beta,kim2018disentangling}.

\noindent
\textbf{Dual-Branch Disentangler Architecture.}
Given a session-level embedding
$\mathbf{x} = \mathcal{E}(S^{[0,T]}) \in \mathbb{R}^D$
from the backbone encoder, we introduce a learnable dual-branch disentangler
$\Phi: \mathbb{R}^D \rightarrow \mathbb{R}^{d_{\mathrm{intent}}} \times \mathbb{R}^{d_{\mathrm{pack}}}$,
which decomposes $\mathbf{x}$ into intent-related and packaging-related latent factors. Here, $d_{\mathrm{intent}}$ and $d_{\mathrm{pack}}$ denote the dimensions of the intent and packaging latent spaces, respectively.

Specifically, $\Phi(\cdot)$ 
 is implemented as a lightweight dual-branch multilayer perceptron~(MLP) on top of the backbone embedding: a shared transformation first extracts common session semantics, followed by two projection heads that map the shared representation into the \emph{intent} and \emph{packaging} subspaces:
\begin{equation}
\mathbf{h} = f_{\mathrm{shared}}(\mathbf{x}), \quad
\mathbf{z}_{\mathrm{intent}} = f_{\mathrm{intent}}(\mathbf{h}), \quad
\mathbf{z}_{\mathrm{pack}} = f_{\mathrm{pack}}(\mathbf{h}),
\end{equation}
producing a pair of latent representations $(\mathbf{z}_{\mathrm{intent}},\mathbf{z}_{\mathrm{pack}})$ for a single live streaming session. For notational brevity, we denote the complete intent branch by $\Phi_{\mathrm{intent}}(\cdot) = f_{\mathrm{intent}} \circ f_{\mathrm{shared}}(\cdot)$ in the subsequent sections.

\noindent
\textbf{Semantic Preservation via Reconstruction.}
To ensure that the disentangled representations jointly preserve sufficient session semantics, we introduce a reconstruction-based regularization.
A decoder $D(\cdot)$ recombines the intent and packaging representations to reconstruct the original session embedding $
\hat{\mathbf{x}} = D(\mathbf{z}_{\mathrm{intent}}, \mathbf{z}_{\mathrm{pack}}),
$ which is implemented as a 2-layer MLP.
The reconstruction loss is defined as
\begin{equation}
\mathcal{L}_{\mathrm{rec}} = \| \mathbf{x} - \hat{\mathbf{x}} \|_2^2,
\end{equation}
where $\|\cdot\|_2$ denotes the Euclidean norm.
This loss prevents degenerate solutions and encourages faithful information preservation across the two latent factors.

\noindent
\textbf{Cross-factor Orthogonality Constraint.}
To further reduce unintended information leakage between intent and packaging representations, we impose a soft orthogonality constraint~\cite{bousmalis2016domain} that penalizes linear correlation between the two latent spaces.
Given a training batch of size $B$, the orthogonality loss is defined as
\begin{equation}
\mathcal{L}_{\mathrm{ortho}} = \frac{1}{B} \left\| \mathbf{Z}_{\mathrm{intent}}^\top \mathbf{Z}_{\mathrm{pack}} \right\|_F^2,
\end{equation}
where $\mathbf{Z}_{\mathrm{intent}}$ and $\mathbf{Z}_{\mathrm{pack}}$ denote the batch-wise matrices of intent and packaging representations, respectively, and $\|\cdot\|_F$ denotes the Frobenius norm.
This regularization acts as a soft constraint that discourages cross-factor entanglement without enforcing strict independence assumptions.

The resulting disentangled latent space provides a structured representation in which stable intent and volatile tactical packaging are explicitly separated.
Next, we introduce counterfactual consistency objectives that operate on this latent factorization to enforce robustness under controlled packaging interventions.

\subsection{Counterfactual Consistency Decoupling}

While latent disentanglement exposes intent- and packaging-related factors, architectural separation alone does not guarantee that the intent representation is invariant to tactical variations.
Under purely observational supervision, intent embeddings may still encode tactic-specific cues that co-occur with malicious behavior in the training data.
To explicitly eliminate such spurious dependencies, inspired by causal intervention~\cite{pearl2009causality,feder2022causal}, we introduce \emph{Counterfactual Consistency Decoupling (CCD)}, which enforces intent invariance under controlled packaging interventions at both the representation and prediction levels.

\subsubsection{Representation-Level CCD}
The representation-level CCD enforces that intent-related representations remain stable when tactical packaging is counterfactually altered.
Otherwise, intent representations are learned only from co-occurring intent–packaging pairs in the training data, and their apparent stability does not imply robustness to unseen tactical realizations.


\noindent
\textbf{Counterfactual Construction.}
Given a training batch of live streaming sessions, we partition samples into \emph{risky} and \emph{safe} groups based on supervision.
To approximate a stable and benign tactical realization, we compute a batch-wise reference packaging representation as the mean of packaging factors from safe sessions, denoted by $\bar{\mathbf{z}}_{\mathrm{pack}}^{\mathrm{safe}}$. The use of the batch-wise mean $\bar{\mathbf{z}}_{\mathrm{pack}}^{\mathrm{safe}}$ serves as a prototypical representation of benign tactical packaging, providing a stable intervention target that is independent of the tactic-specific cues of individual risky sessions.

For each risky session with intent representation $\mathbf{z}_{\mathrm{intent}}^{r}$, we construct a counterfactual session embedding by explicitly intervening on the packaging factor while preserving the intent factor:
$
\mathbf{x}_{\mathrm{CF}}^r = D\left(\mathbf{z}_{\mathrm{intent}}^{r}, \bar{\mathbf{z}}_{\mathrm{pack}}^{\mathrm{safe}}\right),
$
where $D(\cdot)$ denotes the decoder introduced in the disentanglement module.
Similar to counterfactual generation in observational space~\cite{sauercounterfactual}, this operation simulates the same malicious intent expressed under an ordinary, benign packaging.

The counterfactual embedding is then re-encoded by the disentangler to obtain the corresponding counterfactual intent representation:
$
\mathbf{z}_{\mathrm{intent}}^{\mathrm{CF},r} = \Phi_{\mathrm{intent}}\left(\mathbf{x}_{\mathrm{CF}}^r\right).
$

\noindent
\textbf{Latent Consistency Objective.}
To enforce invariance, we adopt a contrastive consistency objective utilizing a triplet-style loss~\cite{schroff2015facenet,chen2020simclr}.
Specifically, the factual intent representation $\mathbf{z}_{\mathrm{intent}}^{r}$ serves as the \emph{anchor}.
Its counterfactual counterpart $\mathbf{z}_{\mathrm{intent}}^{\mathrm{CF},r}$, obtained by intervening on narrative packaging while preserving intent, is treated as the \emph{positive},
while intent representations from safe sessions act as \emph{negatives}.
The representation-level CCD loss is defined as:
\begin{equation}
\mathcal{L}_{\mathrm{CCD}}^{\mathrm{rep}} =
\max \left(
0,
m
+ \mathbb{E}_{r,s}\!\left[\mathrm{Sim}\!\left(\mathbf{z}_{\mathrm{intent}}^{r}, \mathbf{z}_{\mathrm{intent}}^{s}\right)\right]
- \mathrm{Sim}\!\left(\mathbf{z}_{\mathrm{intent}}^{r}, \mathbf{z}_{\mathrm{intent}}^{\mathrm{CF},r}\right)
\right),
\end{equation}
where $\mathbf{z}_{\mathrm{intent}}^{s}$ denotes intent representations from all safe sessions in the batch.
$\mathrm{Sim}(\cdot,\cdot)$ denotes cosine similarity, and $m$ is a margin hyperparameter.
This objective encourages intent representations to remain invariant under counterfactual packaging while maintaining separation from benign intent patterns.

\noindent
\textbf{Gradient Blocking Strategy.} 
In our implementation, during the computation of $\mathcal{L}_{\mathrm{CCD}}^{\mathrm{rep}}$, we block the gradient flow through the counterfactual generation process (i.e., the decoder $D$ and the re-disentanglement of $\textbf{x}_{\mathrm{CF}}$). This ensures that the loss specifically optimizes the disentangler $\Phi$ to map the counterfactual input back to its original intent manifold, rather than implicitly shifting the counterfactual construction itself to simplify the task.

\subsubsection{Prediction-Level CCD}

While representation-level CCD constrains the latent space, it does not directly prevent the downstream classifier from exploiting residual tactic-related cues.
Hence, similar to~\cite{veitch2021counterfactual}, prediction-level CCD should enforce causal consistency at the decision level, requiring the risk predictor to produce stable outputs under packaging interventions.

The core intuition is that if the disentanglement is successful, replacing a risky session's original packaging $\mathbf{z}_{\mathrm{pack}}^{r}$ with a safe reference $\bar{\mathbf{z}}_{\mathrm{pack}}^{\mathrm{safe}}$ should not alter its risk nature. Therefore, the predictor's output for the counterfactual session (which carries the same malicious intent but is re-wrapped in a benign style) should remain consistent with the factual prediction.

\noindent
\textbf{Predictive Consistency Objective.}
For a risky session, we compute the factual and counterfactual \emph{logits} using the same intent representation:
\begin{equation}
\ell = g\!\left(\mathbf{z}_{\mathrm{intent}}^{r}\oplus\mathbf{z}_{\mathrm{pack}}^{r}\right),\quad
\ell_{\mathrm{CF}} = g\!\left(\mathbf{z}_{\mathrm{intent}}^{r}\oplus \bar{\mathbf{z}}_{\mathrm{pack}}^{\mathrm{safe}}\right),
\end{equation}
where $g(\cdot)$ denotes the risk predictor before activation and $\oplus$ denotes concatenation.
To enforce predictive invariance under counterfactual packaging intervention, we minimize the discrepancy between the two logits:
\begin{equation}
\mathcal{L}_{\mathrm{CCD}}^{\mathrm{pred}} =
\left\| \ell - \ell_{\mathrm{CF}} \right\|_2^2.
\end{equation}
Unlike representation-level CCD, 
$\mathcal{L}_{\mathrm{CCD}}^{\mathrm{pred}}$ allows end-to-end gradient propagation, explicitly discouraging reliance on tactic-induced shortcuts.

Together, the two levels of CCD form a two-stage causal regularization mechanism.
Representation-level CCD enforces invariance in the latent intent space, while prediction-level CCD ensures that such invariance is respected by the decision function.
By enabling representation-level invariance and predictive consistency, LPCD establishes a robust causal bridge from latent factorization to final risk assessment, ensuring that the decision boundary is inherently resilient to the ``chameleon-like'' evolution of adversarial packaging.

\subsection{Risk Prediction and Training Objective}

In the following, we formulate the joint optimization objective of the plug-in LPCD framework.

 \noindent \textbf{Main Risk Prediction.}
To produce the final risk score, we employ the risk predictor $g(\cdot)$ that takes the disentangled factors as input. 
To capture the full session context while emphasizing the disentangled structure, we concatenate the intent and packaging representations as the final feature vector: $
\hat{y} = \mathrm{Sigmoid}\!\left(g(\mathbf{z}_{\mathrm{intent}}\oplus\mathbf{z}_{\mathrm{pack}})\right),
$
where $\hat{y}\in(0,1)$ denotes the predicted risk probability.
The primary objective is to minimize the binary cross-entropy (BCE) loss under standard supervision:
\begin{equation}
\mathcal{L}_{\mathrm{main}} = - \frac{1}{B} \sum_{i=1}^B
\left[
y_i \log \hat{y}_i + (1-y_i) \log(1-\hat{y}_i)
\right],
\end{equation}
where $y_i \in \{0, 1\}$ denotes the ground-truth risk label.

\noindent
\textbf{Joint Optimization Objective.}
LPCD is trained end-to-end by simultaneously optimizing the predictive performance and the constraints of the latent space. The total loss function is defined as a weighted combination of all previously introduced objectives:
\begin{equation}
    \mathcal{L}_{\text{total}} = \mathcal{L}_{\mathrm{main}} + \lambda_{\mathrm{rec}} \mathcal{L}_{\mathrm{rec}} + \lambda_{\mathrm{ortho}} \mathcal{L}_{\mathrm{ortho}} + \lambda_{\mathrm{CCD}}^{\mathrm{rep}} \mathcal{L}_{\mathrm{CCD}}^{\mathrm{rep}} + \lambda_{\mathrm{CCD}}^{\mathrm{pred}}\mathcal{L}_{\mathrm{CCD}}^{\mathrm{pred}},
\end{equation}
where $\lambda_{\mathrm{rec}}, \lambda_{\mathrm{ortho}} , \lambda _{\mathrm{CCD}}^{\mathrm{rep}}, \lambda_{\mathrm{CCD}}^{\mathrm{pred}}$ are hyperparameters that balance the trade-off between semantic preservation, factor orthogonality, and dual-level causal consistency. This joint supervision prevents the model from exploiting spurious correlations, ensuring the decision boundary is anchored on stable intent-related factors.

\subsection{Post-hoc Magnitude Calibration at Inference}
\label{sec:met:cal}
While the CCD module enforces semantic invariance during training, adversarial attackers may still induce \emph{tactical magnitude shifts} in the packaging manifold during deployment.
Such shifts manifest as changes in the latent energy of $\mathbf{z}_{\mathrm{pack}}$, which can destabilize the predictor even when the underlying semantic content remains unchanged.
Inspired by test-time normalization techniques~\cite {li2018adaptive}, to ensure robust deployment under evolving tactics, we introduce a lightweight post-hoc calibration mechanism that rectifies test-time packaging magnitudes using training-stage statistics.

\noindent
\textbf{Online Magnitude Tracking.}
To handle the high variance of live streaming traffic, we maintain a running estimate of the second-order statistics of the packaging representation.
Let $\sigma_{\mathrm{train}, d}$ denote the Root Mean Square (RMS) of the $d$-th dimension of $\mathbf{z}_{\mathrm{pack}}$ computed over the safe samples from the training set.
During inference, we estimate the test-stage magnitude using a sliding batch of incoming sessions.
Specifically, given a mini-batch $\mathcal{B}^{(t)}$ at inference step $t$, the test-time RMS is updated as:
\begin{equation}
\sigma_{\mathrm{test}, d}^{(t)} =
(1 - \alpha)\, \sigma_{\mathrm{test}, d}^{(t-1)}
+ \alpha \sqrt{
\frac{1}{|\mathcal{B}^{(t)}|}
\sum_{\mathbf{z} \in \mathcal{B}^{(t)}} (\mathbf{z}_{\mathrm{pack}, d})^2
},
\end{equation}
where $\alpha \in (0,1]$ is a momentum coefficient.
The tracking process is initialized with $\sigma_{\mathrm{test}, d}^{(0)}=\sigma_{\mathrm{train}, d}$.
Note that in offline evaluation, we approximate the online update by computing $\sigma_{\mathrm{test}, d}$ from the current test mini-batch only.

\noindent
\textbf{Magnitude Rectification.}
Based on the tracked statistics, we construct a diagonal calibration matrix
$\boldsymbol{\Gamma}^{(t)} \in \mathbb{R}^{d_{\mathrm{pack}}\times d_{\mathrm{pack}}}$
to rescale the packaging representation:
\begin{equation}
\boldsymbol{\Gamma}^{(t)} = \mathrm{diag}\!\left(
\gamma_1^{(t)}, \ldots, \gamma_{d_{\mathrm{pack}}}^{(t)}
\right),
\quad
\gamma_d^{(t)} = \frac{\sigma_{\mathrm{train}, d}}{\sigma_{\mathrm{test}, d}^{(t)}}.
\end{equation}
The calibrated packaging representation is then obtained via a simple diagonal transformation: $
\tilde{\mathbf{z}}_{\mathrm{pack}} =
\boldsymbol{\Gamma}^{(t)} \mathbf{z}_{\mathrm{pack}}.$
The final calibrated risk score is produced as:
$
\hat{y}_{\mathrm{cal}} =
\mathrm{Sigmoid}\!\left(
g(\mathbf{z}_{\mathrm{intent}}\oplus \tilde{\mathbf{z}}_{\mathrm{pack}})
\right).
$

By aligning the latent energy of the packaging factor to training-stage statistics, this calibration module mitigates tactic-induced magnitude perturbations at inference time.
Importantly, this calibration operates purely at the \emph{statistical level}.
It introduces no additional learnable parameters, requires no gradient-based optimization, and incurs only negligible inference-time overhead, making it suitable for high-throughput live streaming scenarios.
\section{Experiments}
\begin{table}[t]
\caption{Statistics of the May and June datasets.}
\label{table:exp:dataset}
\centering
\resizebox{0.95\linewidth}{!}{
  \begin{tabular}{c|c|cccc} 
  \toprule
   &  & \#Sessions & \#Avg.Actions & \#Avg.Users & Avg.Time~(min) \\ 
  \midrule
  \multirow{4}{*}{\textbf{May}} 
   & train & $176{,}347$ & 709 & 35 & 30.0 \\ 
   & val & $23{,}562$ & 704 & 36 & 29.6 \\ 
   & ID test & $22{,}462$ & 740 & 37 & 29.7 \\ 
   & OOD test & $\textit{15,320}$ & $\textit{666}$ & $\textit{44}$ & $\textit{28.5}$ \\ 
  \midrule
  \multirow{4}{*}{\textbf{June}} 
   & train & $79{,}552$ & 700 & 36 & 30.0 \\ 
   & val & $10{,}934$ & 767 & 40 & 29.1 \\ 
   & ID test & $10{,}967$ & 725 & 37 & 29.1 \\ 
   & OOD test & $\textit{16,722}$ & $\textit{679}$ & $\textit{44}$ & $\textit{28.6}$\\ 
  \bottomrule
  \end{tabular}
}
\end{table}

\begin{table*}[t]
\scriptsize
\centering
\setlength{\tabcolsep}{1pt} 
\caption{Overall Performance Comparison on May and June Datasets. Metrics: PR-AUC (AUC), F1-score (F1), R@0.1FPR (R.1), and FPR@0.9R (FPR.9). Best and second-best results are in \textbf{bold} and shaded red, underlined and shaded orange, respectively; backbone SOTA is in \textbf{bold} and shaded green. `$^{\ast}$' indicates $p < 0.05$.}
\label{tab:exp:main}

\resizebox{\textwidth}{!}{
\begin{tabular}{c|c|cccc|cccc||cccc|cccc}
\toprule
\multicolumn{2}{c|}{\multirow{3}{*}{\textbf{Methods}}} & \multicolumn{8}{c||}{\textbf{Trained on May (05/20--06/03)}} & \multicolumn{8}{c}{\textbf{Trained on June (06/04--06/10)}} \\
\cmidrule(lr){3-10} \cmidrule(lr){11-18}
\multicolumn{2}{c|}{} & \multicolumn{4}{c|}{\textbf{May ID Test Set (06/13--06/14)}} & \multicolumn{4}{c||}{\textbf{May OOD Test Set (09/23--09/24)}} & \multicolumn{4}{c|}{\textbf{June ID Test Set (06/16)}} & \multicolumn{4}{c}{\textbf{June OOD Test Set (10/16--10/17)}} \\
\cmidrule(lr){3-6} \cmidrule(lr){7-10} \cmidrule(lr){11-14} \cmidrule(lr){15-18}
\multicolumn{2}{c|}{}  & \tiny{AUC}$\uparrow$ & \tiny{F1}$\uparrow$ & \tiny{R.1}$\uparrow$ & \tiny{FPR.9}$\downarrow$ & \tiny{AUC}$\uparrow$ & \tiny{F1}$\uparrow$ & \tiny{R.1}$\uparrow$ & \tiny{FPR.9}$\downarrow$ & \tiny{AUC}$\uparrow$ & \tiny{F1}$\uparrow$ & \tiny{R.1}$\uparrow$ & \tiny{FPR.9}$\downarrow$ & \tiny{AUC}$\uparrow$ & \tiny{F1}$\uparrow$ & \tiny{R.1}$\uparrow$ & \tiny{FPR.9}$\downarrow$ \\
\midrule

\multicolumn{18}{c}{\cellcolor{gray!20}\textit{\textbf{Backbones}}} \\

\midrule
\multirow{3}{*}{\shortstack{\textit{Sequence}\\ \textit{Models}}} 
& Transformer & 0.7189 & 0.6668 & 0.8394 & 0.1580 & 0.6728 & 0.6007 & 0.7978 & 0.2008 & 0.6801 & 0.6341 & 0.8225 & 0.1565 & 0.6208 & 0.5907 & 0.7636 & 0.2545 \\
& Reformer    & 0.7293 & 0.6752 & 0.8575 & 0.1436 & 0.6570 & 0.5842 & 0.7890 & 0.2126 & 0.6911 & 0.6395 & 0.8104 & 0.1760 & 0.6189 & 0.5967 & 0.7562 & 0.2638 \\
& Informer    & 0.7246 & 0.6708 & 0.8438 & 0.1555 & 0.6586 & 0.6007 & 0.7949 & 0.2232 & 0.6879 & 0.6391 & 0.8375 & 0.1601 & 0.6028 & 0.5902 & 0.7508 & 0.2661 \\
\midrule[0.1pt]
\multirow{4}{*}{\shortstack{\textit{MIL}\\ \textit{Methods}}}
& MIL-LET     & 0.7241 & 0.6749 & 0.8546 & 0.1418 & 0.6643 & 0.5920 & 0.7978 & 0.1932 & 0.6942 & 0.6528 & 0.8455 & 0.1499 & 0.6050 & 0.5191 & 0.7676 & 0.2741 \\
& TimeMIL     & 0.7353 & 0.6790 & 0.8599 & 0.1436 & 0.6443 & 0.5864 & 0.7816 & 0.1904 & 0.6963 & 0.6471 & 0.8495 & 0.1367 & 0.6316 & 0.5983 & 0.7763 & 0.2288 \\
& TAIL-MIL    & 0.7316 & 0.6785 & 0.8570 & 0.1341 & 0.6606 & 0.5793 & 0.7904 & 0.2008 & 0.7029 & 0.6509 & 0.8205 & 0.1555 & 0.6365 & 0.5869 & 0.7776 & 0.2391 \\
& \cellcolor{green!6}\textbf{AC-MIL} & \cellcolor{green!6}\textbf{0.7676} & \cellcolor{green!6}\textbf{0.7002} & \cellcolor{green!6}\textbf{0.8722} & \cellcolor{green!6}\textbf{0.1260} & \cellcolor{green!6}\textbf{0.7045} & \cellcolor{green!6}\textbf{0.6428} & \cellcolor{green!6}\textbf{0.8118} & \cellcolor{green!6}\textbf{0.1714} & \cellcolor{green!6}\textbf{0.7311} & \cellcolor{green!6}\textbf{0.6777} & \cellcolor{green!6}\textbf{0.8546} & \cellcolor{green!6}\textbf{0.1345} & \cellcolor{green!6}\textbf{0.6858} & \cellcolor{green!6}\textbf{0.6235} & \cellcolor{green!6}\textbf{0.7957} & \cellcolor{green!6}\textbf{0.2130} \\
\midrule
\multicolumn{18}{c}{\cellcolor{gray!20}\textit{\textbf{Best Backbone~(AC-MIL) + OOD Plug-ins}}} \\
\midrule
\multirow{3}{*}{\textit{IL}}
& + IRM       & 0.7699 & 0.7033 & 0.8781 & 0.1213 & 0.7098 & 0.6408 & 0.8213 & 0.1769 & 0.7317 & \cellcolor{orange!12}\underline{0.6836} & 0.8537 & 0.1403 & 0.6905 & 0.6244 & 0.7991 & 0.2162 \\
& + VREx      & 0.7626 & 0.6969 & 0.8707 & 0.1303 & 0.6999 & 0.6330 & 0.8125 & 0.1836 & 0.7307 & 0.6744 & 0.8566 & 0.1384 & 0.6852 & 0.6150 & \cellcolor{orange!12}\underline{0.8058} & 0.2226 \\
& + IB-IRM    & 0.7719 & \cellcolor{orange!12}\underline{0.7080} & 0.8766 & 0.1219 & 0.7103 & 0.6407 & 0.8140 & 0.1783 & 0.7286 & 0.6757 & 0.8556 & 0.1422 & 0.6849 & 0.6260 & 0.7950 & 0.2144 \\
\midrule[0.1pt]
\multirow{2}{*}{\textit{DA}}
& + MIXUP     & 0.7726 & 0.7018 & 0.8776 & 0.1211 & 0.7062 & 0.6442 & 0.8257 & 0.1780 & 0.7279 & 0.6752 & 0.8445 & 0.1421 & 0.6851 & 0.6277 & 0.7964 & \cellcolor{orange!12}\underline{0.2000} \\
& + CORAL     & 0.7676 & 0.7029 & 0.8692 & 0.1315 & 0.7070 & 0.6378 & 0.8184 & 0.1767 & 0.7327 & 0.6794 & 0.8602 & 0.1313 & \cellcolor{orange!12}\underline{0.6940} & 0.6221 & 0.8051 & 0.2206 \\
\midrule[0.1pt]
\multirow{2}{*}{\textit{DRO}}
& + GroupDRO  & 0.7716 & 0.7049 & 0.8771 & 0.1205 & 0.7127 & 0.6446 & 0.8191 & 0.1789 & 0.7294 & 0.6781 & 0.8538 & 0.1404 & 0.6873 & 0.6241 & 0.7971 & 0.2162 \\
& + ASGDRO    & 0.7715 & 0.7038 & 0.8766 & 0.1222 & \cellcolor{orange!12}\underline{0.7144} & 0.6443 & \cellcolor{orange!12}\underline{0.8235} & 0.1773 & 0.7335 & 0.6811 & 0.8455 & 0.1400 & 0.6884 & 0.6249 & 0.7984 & 0.2197 \\
\midrule[0.1pt]
\multirow{2}{*}{\textit{EI}}
& + EIIL      & 0.7686 & 0.6824 & 0.8756 & 0.1207 & 0.7076 & 0.6409 & 0.8169 & 0.1743 & \cellcolor{orange!12}\underline{0.7375} & 0.6601 & \cellcolor{orange!12}\underline{0.8636} & \cellcolor{orange!12}\underline{0.1299} & 0.6877 & 0.6170 & 0.7971 & 0.2229 \\
& + FOIL      & \cellcolor{orange!12}\underline{0.7747} & 0.7012 & \cellcolor{orange!12}\underline{0.8790} & \cellcolor{orange!12}\underline{0.1191} & 0.7097 & \cellcolor{orange!12}\underline{0.6463} & 0.8191 & \cellcolor{orange!12}\underline{0.1713} & 0.7334 & 0.6760 & \cellcolor{orange!12}\underline{0.8636} & 0.1314 & 0.6828 & \cellcolor{orange!12}\underline{0.6286} & 0.8031 & 0.2111 \\
\midrule[0.1pt]
\multicolumn{2}{c|}{\cellcolor{red!16}\textbf{+~LPCD~(Ours)}} & \cellcolor{red!16}\textbf{0.7841*} & \cellcolor{red!16}\textbf{0.7121*} & \cellcolor{red!16}\textbf{0.8832*} & \cellcolor{red!16}\textbf{0.1158*} & \cellcolor{red!16}\textbf{0.7300*} & \cellcolor{red!16}\textbf{0.6828*} & \cellcolor{red!16}\textbf{0.8529*} & \cellcolor{red!16}\textbf{0.1589*} & \cellcolor{red!16}\textbf{0.7454*} & \cellcolor{red!16}\textbf{0.6877*} & \cellcolor{red!16}\textbf{0.8768*} & \cellcolor{red!16}\textbf{0.1292*} & \cellcolor{red!16}\textbf{0.7287*} & \cellcolor{red!16}\textbf{0.6779*} & \cellcolor{red!16}\textbf{0.8600*} & \cellcolor{red!16}\textbf{0.1732*} \\ 
\midrule[0.1pt]
\multicolumn{2}{c|}{\textit{\textbf{Gain over AC-MIL}}} & \textit{+2.1\%} & \textit{+1.7\%} & \textit{+1.3\%} & \textit{-8.1\%} & \textit{+3.6\%} & \textit{+6.2\%} & \textit{+5.1\%} & \textit{-7.3\%} & \textit{+2.0\%} & \textit{+1.5\%} & \textit{+2.6\%} & \textit{-4.0\%} & \textit{+6.3\%} & \textit{+8.7\%} & \textit{+2.1\%} & \textit{-18.7\%} \\ 
\multicolumn{2}{c|}{\textit{\textbf{Gain over Best Plug-in}}} & \textit{+1.2\%} & \textit{+0.6\%} & \textit{+0.5\%} & \textit{-2.8\%} & \textit{+2.2\%} & \textit{+5.6\%} & \textit{+3.6\%} & \textit{-7.2\%} & \textit{+1.1\%} & \textit{+1.0\%} & \textit{+1.5\%} & \textit{-1.0\%} & \textit{+5.0\%} & \textit{+7.8\%} & \textit{+5.4\%} & \textit{-13.4\%} \\ 

\bottomrule
\end{tabular}
}
\end{table*}

In this section, we evaluate LPCD on large-scale industrial data to answer the following research questions:

\begin{itemize}[leftmargin=*,topsep=5pt]
    \item \textbf{RQ1}: Does LPCD outperform strong baselines under both in-distribution and tactical OOD settings?
    \item \textbf{RQ2}: What is the contribution of each component in LPCD?
    \item \textbf{RQ3}: How does LPCD compare with a retraining oracle in terms of performance and efficiency?
    \item \textbf{RQ4}: Does LPCD disentangle intent-invariant risk signals from tactical packaging variations in the latent space?
    \item \textbf{RQ5}: Can LPCD be effectively applied as a plug-in to different backbone models?
    \item \textbf{RQ6}: Does LPCD improve performance in online deployment?
\end{itemize}

\subsection{Experimental Setup}

\subsubsection{Datasets}
We collect two large-scale industrial live-streaming datasets from the Douyin Live-streaming platform\footnote{All data were collected and processed in compliance with the platform’s privacy policy.}, denoted as \textbf{May} and \textbf{June}~\footnote{\url{https://huggingface.co/datasets/ByteDance/LiveStreamingRiskControl}}.
To assess robustness against tactical evolution, each dataset is temporally partitioned into \emph{training}, \emph{validation}, \emph{in-distribution (ID) test}, and a \emph{tactical OOD test} set.
For the \textbf{May} dataset, training data spans 05/20/2025--06/03/2025, followed by a validation set from 06/11/2025 to 06/12/2025, an ID test set on 06/13/2025--06/14/2025, and an OOD test set spans from 09/23/2025 to 09/24/2025.
The \textbf{June} dataset uses 06/04/2025--06/10/2025 for training, 06/15/2025 for validation, and 06/16/2025 as the ID test set, with its OOD evaluation on 10/16/2025--10/17/2025. 
Table~\ref{table:exp:dataset} presents the basic statistics of our datasets.

\textit{Action Space and Modalities.}
Sessions are represented by heterogeneous action sequences involving both hosts and viewers. Viewer-side actions include entries, comments (danmaku), virtual gifting, and social interactions (i.e., likes, shares, co-stream
requests, and group joins). In addition to the start of the stream, host-side signals provide semantic context through speech transcripts obtained via ASR and on-screen text extracted by OCR. Textual content is encoded using a Chinese-BERT encoder\footnote{https://huggingface.co/google-bert/bert-base-chinese}. 

\textit{Session Processing.}
Following prior work~\cite{qiao2026livelieactionawarecapsule}, each live streaming session is truncated to its first 30 minutes to reflect early-stage risk detection. To focus on high-impact interactions, we retain signals from the top 50 most active viewers per session. Following industrial risk control practice, all malicious sessions are preserved, while benign sessions are down-sampled to maintain a 1:10 class ratio.

\subsubsection{Baselines.}
\emph{(a) Backbones.}
Following prior practice~\cite{qiao2026livelieactionawarecapsule}, we consider two families of backbones as candidates:
\textit{\textbf{Sequence models}} including Transformer~\cite{vaswani2017attention}, Reformer~\cite{kitaev2020reformer}, and Informer~\cite{zhou2021informer}; and \textit{\textbf{Multiple Instance Learning (MIL) methods}} including MIL-LET~\cite{early2024inherently}, TimeMIL~\cite{chen2024timemil}, TAIL-MIL~\cite{jang2025tail}, and the SOTA AC-MIL~\cite{qiao2026livelieactionawarecapsule}.

\emph{(b) OOD Plug-ins.}
We compare LPCD with representative OOD generalization plug-ins from four paradigms:
\textit{\textbf{Invariant Learning (IL)}}, including IRM~\cite{arjovsky2019invariant},
VREx~\cite{krueger2021out},
and IB-IRM~\cite{ahuja2021invariance};
\textit{\textbf{Data Augmentation and Alignment (DA)}}, including Mixup~\cite{yan2020improve}
and CORAL~\cite{sun2016deep};
\textit{\textbf{Distributionally Robust Optimization (DRO)}}, including GroupDRO~\cite{Sagawa*2020Distributionally}
and ASGDRO~\cite{kim2025sufficient};
and
\textit{\textbf{Environment Inference (EI)}}, including EIIL~\cite{creager2021environment}
and FOIL~\cite{liu2024time}.
Note that more baseline details can be found in Appendix~\ref{sec:app:exp}.

\subsubsection{Implementation Details.}
All the models are trained using AdamW~\cite{loshchilov2018decoupled} with a learning rate and weight decay of $1\mathrm{e}{-4}$. 
The session embedding dimension is set to 128, while disentangled representations $\mathbf{z}_{\mathrm{intent}}$ and $\mathbf{z}_{\mathrm{pack}}$ are both 32-dimensional. 
The causal consistency loss weights $\lambda_{\mathrm{CCD}}^{\mathrm{rep}}$ and $\lambda_{\mathrm{CCD}}^{\mathrm{pred}}$ are selected via grid search over $\{0.5,1.0,2.0\}$ and $\{0.05,0.1,0.2,0.5,1.0\}$, respectively. Hyperparameter sensitivity results are provided in Appendix~\ref{sec:app:hyper}.

Models are trained for up to 100 epochs with a batch size of 128 and an early stopping patience of 20. 
To stabilize optimization, only the primary BCE loss $\mathcal{L}_{\mathrm{main}}$ is optimized during the first 5 warm-up epochs. 
Following AC-MIL~\cite{qiao2026livelieactionawarecapsule}, all backbone architectures use a dropout rate of 0.1. 
The margin hyperparameter $m$ and momentum coefficient $\alpha$ are fixed at 1.0 and 0.1, respectively. 
We set $\lambda_{\mathrm{rec}}=1.0$, while $\lambda_{\mathrm{ortho}}$ is set to $5\mathrm{e}{-4}$ for May and $1\mathrm{e}{-3}$ for June.

\subsubsection{Evaluation Metrics.}
In all experiments, we report \textbf{PR-AUC}, \textbf{F1-score}, \textbf{R@0.1FPR}, and \textbf{FPR@0.9R}. 
PR-AUC and F1-score assess performance under class imbalance, where PR-AUC is preferred over ROC-AUC for its sensitivity to positive cases. 
R@0.1FPR reports recall at a fixed false positive rate of 10\%, while FPR@0.9R measures the false positive rate at 90\% recall. 
These threshold-based metrics align with practical moderation requirements by balancing high-risk coverage and false alarm control.

\subsection{Overall Performance~(RQ1)}
Table~\ref{tab:exp:main} reports the overall performance on the \textbf{May} and \textbf{June} datasets, covering both ID and OOD evaluation settings. We summarize four key observations.

\textit{\textbf{LPCD consistently outperforms all baselines across datasets and distribution settings.}}
Across both the May and June datasets, LPCD consistently outperforms all baselines on all four metrics under both ID and OOD test settings. These gains hold across different temporal splits and evaluation criteria, indicating that LPCD provides a stable and general performance improvement. 

\textit{\textbf{LPCD exhibits amplified advantages under tactical OOD shifts.}}
We observe a universal performance degradation for all models as the temporal gap increases; e.g., the PR-AUC of AC-MIL drops by 6.2\%--8.2\% when transitioning to OOD sets. However, LPCD's relative advantages become markedly more pronounced in these challenging scenarios. On the May OOD set, LPCD improves PR-AUC by 3.6\% over AC-MIL and 2.2\% over the strongest OOD plug-in, with even larger relative gains on F1-score (+6.2\%). This widening gap directly supports our claim that LPCD is uniquely effective under \textit{tactical OOD} conditions.

\textit{\textbf{LPCD surpasses generic OOD plug-ins through specialized causal intervention.}}
LPCD notably outperforms a wide spectrum of OOD techniques with the same backbone. While these baselines aim to improve robustness via generic regularization or implicit environment inference, LPCD explicitly intervenes on latent narrative packaging to enforce counterfactual consistency. The persistent performance gap indicates that LPCD captures complementary causal structures that generic OOD heuristics fail to model.

\textit{\textbf{LPCD delivers superior recall--false-alarm trade-offs for real-world moderation.}}
Beyond aggregate metrics, LPCD achieves consistent improvements on threshold-sensitive indicators critical to industrial systems. Across both datasets, LPCD increases R@0.1FPR while simultaneously reducing FPR@0.9R. Notably, the 18.7\% relative reduction in FPR@0.9R on the June OOD set demonstrates LPCD's ability to substantially reduce moderation burden under severe tactical shifts.

\subsection{Ablation Study~(RQ2)}
\begin{table}[t]
\centering
\caption{Ablation results on June OOD test set. $\mathcal{L}_\mathrm{dis} = \{\mathcal{L}_\mathrm{rec}, \mathcal{L}_\mathrm{ortho}\}$ and $\mathcal{L}_\mathrm{ccd} = \{\mathcal{L}_\mathrm{CCD}^\mathrm{rep}, \mathcal{L}_\mathrm{CCD}^\mathrm{pred}\}$. TT-Calibration refers to Post-hoc Magnitude Calibration at inference.}
\label{tab:exp:ablation}
\resizebox{0.48\textwidth}{!}{ 
\begin{tabular}{l|cccc}
\toprule
 \multirow{2}{*}{\textbf{Variants}} & \multicolumn{4}{c}{\textbf{June OOD Test Set}} \\
\cmidrule(lr){2-5} 
& PR-AUC$\uparrow$ & F1-score$\uparrow$
& R@0.1FPR$\uparrow$ & FPR@0.9R$\downarrow$ \\
\midrule
 Backbone (AC-MIL) & 0.6858 & 0.6235 & 0.7957 & 0.2130\\
\midrule
LPCD w/o $\mathcal{L}_\mathrm{dis}$ (Only $\mathcal{L}_\mathrm{ccd}$) &0.6881  &0.6236  &0.7957  & 0.2117\\
LPCD w/o $\mathcal{L}_\mathrm{ccd}$ (Only $\mathcal{L}_\mathrm{dis}$) &0.6889  &0.6311  &0.7977  &0.2269  \\
LPCD w/o $\mathcal{L}_\mathrm{rec}$ &0.6812  & 0.6207 &0.7910  &0.2193 \\
LPCD w/o $\mathcal{L}_\mathrm{ortho}$ &0.6853  & 0.6240 & 0.7883 & 0.2179 \\
LPCD w/o $\mathcal{L}_\mathrm{CCD}^\mathrm{rep}$ 
&0.6945  &0.6393  &0.8064  &0.2179 \\
LPCD w/o $\mathcal{L}_\mathrm{CCD}^\mathrm{pred}$ &0.6929  &0.6357  &0.8024  &0.2132  \\
\midrule
LPCD w/o TT-Calibration & 0.7053 &0.6388  & 0.8178 & 0.2041 \\
\midrule
\rowcolor{gray!10}\textbf{LPCD} & \textbf{0.7287} & \textbf{0.6779} & \textbf{0.8600} & \textbf{0.1732} \\
\bottomrule
\end{tabular}
}
\end{table}

To analyze the contribution of each component in LPCD, we conduct an ablation study on the June OOD test set, as shown in Table~\ref{tab:exp:ablation}.
More ablation results on test-time calibration can be found in Appendix~\ref{sec:app:ab}.

\noindent
\textbf{Decoupling and intervention are mutually dependent.}
Removing either the disentanglement losses ($\mathcal{L}_{\mathrm{dis}}$) or the counterfactual losses ($\mathcal{L}_{\mathrm{ccd}}$) yields only marginal improvements over the AC-MIL backbone.
This indicates that effective intervention relies on explicitly decoupled representations, while decoupling alone is insufficient without counterfactual supervision.

\noindent
\textbf{Partial decoupling is detrimental.}
Removing a single decoupling constraint ($\mathcal{L}_{\mathrm{rec}}$ or $\mathcal{L}_{\mathrm{ortho}}$) causes a larger performance drop than removing both.
This suggests that inconsistent decoupling introduces a harmful inductive bias, whereas removing both allows the model to fall back to a stable but non-causal representation.

\noindent
\textbf{Both representation- and prediction-level CCD are required.}
Ablating either $\mathcal{L}_{\mathrm{CCD}}^{\mathrm{rep}}$ or $\mathcal{L}_{\mathrm{CCD}}^{\mathrm{pred}}$ consistently degrades performance, confirming that robustness to tactical shifts must be enforced at both the latent representation and final decision stages.

\noindent
\textbf{Test-time calibration matters.}
Removing test-time calibration significantly reduces PR-AUC (from \textbf{0.7287} to \textbf{0.7053}), showing that calibration serves as an effective last-mile adjustment for residual packaging shifts at inference time.

\subsection{Efficiency Study~(RQ3)}

\begin{table}[t]
\centering
\caption{
Efficiency comparison between LPCD and a \textbf{Retraining Oracle} on the June OOD test set (10/16--10/17).
Retraining cost and inference latency are reported as wall-clock time measured in offline experiments. Inference latency is averaged over three runs on the full test set~(16,722 samples). Metrics: PR-AUC (AUC), F1-score (F1), R@0.1FPR (R.1), and FPR@0.9R (FPR.9).
}

\label{tab:exp:oracle}
\resizebox{0.48\textwidth}{!}{
\begin{tabular}{l|cccc|cc}
\toprule
\multirow{2}{*}{\textbf{Method}} 
& \multicolumn{4}{c|}{\textbf{Performance}} 
& \multicolumn{2}{c}{\textbf{Operational Cost}} \\
\cmidrule(lr){2-5} \cmidrule(lr){6-7}
& AUC$\uparrow$ & F1$\uparrow$ & R.1$\uparrow$ & FPR.9$\downarrow$ 
& Retrain Time & Inf. Latency \\
\midrule
AC-MIL (Fixed) 
& 0.6858 & 0.6235 & 0.7957 & 0.2130 
& -- & 714~s \\

AC-MIL (Oracle) 
& \textbf{0.7303} & \underline{0.6603} & \underline{0.8231} & \underline{0.2016} 
& \textbf{21.8 h} & 717~s \\

\midrule
\rowcolor{gray!10}
\textbf{LPCD (Fixed)} 
& \underline{0.7287} & \textbf{0.6779} & \textbf{0.8600} & \textbf{0.1732} 
& -- & \textbf{654~s} \\
\bottomrule
\end{tabular}
}
\end{table}
To evaluate efficiency under label latency, we compare LPCD on June OOD test set~(10/16--10/17) with a \emph{Retraining Oracle} that fully retrains the backbone using the latest labeled data.
The oracle is retrained on data from 10/08--10/14 with validation on 10/15, while LPCD is applied to a fixed model trained four months earlier (06/04--06/10), without any parameter updates. 

As shown in Table~\ref{tab:exp:oracle}, LPCD achieves performance comparable to the retraining oracle with zero retraining cost.
Although the oracle slightly outperforms LPCD on PR-AUC, LPCD consistently performs better on all operational metrics (F1, R@0.1FPR, and FPR@0.9R).
This indicates that LPCD improves decision quality under strict operating constraints, rather than merely adapting to recent class prevalence.
Moreover, LPCD reduces inference latency.
This benefit comes from its decoupled heads operating on compact intent and packaging representations ($32+32$ dimensions), instead of the high-dimensional ($128$) backbone features.
Overall, LPCD provides a robust and efficient alternative to frequent retraining for real-time risk detection.

\subsection{Case Study~(RQ4)}
\begin{figure}[b]
  \centering
\includegraphics[width=0.45\textwidth]{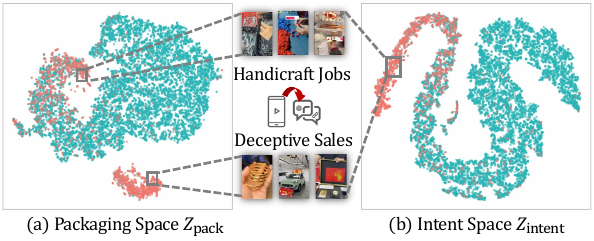}
  \caption{t-SNE visualization of decoupled representations.
Packaging representations separate sessions by surface tactics, while intent representations align sessions sharing the same underlying malicious objective.}
    \vspace{-1em}
  \label{fig:exp:tsne}
\end{figure}
To examine the effect of causal decoupling, we present a case study on two prevalent deceptive tactics: 
\textit{Handicraft Jobs} (fake home-based work recruitment) and 
\textit{Deceptive Sales} (luxury goods offered at extremely low prices).
As shown in Figure~\ref{fig:exp:tsne}(a), these sessions form well-separated clusters in the \textit{Packaging Space}, reflecting their distinct surface presentations.
In contrast, Figure~\ref{fig:exp:tsne}(b) shows that the same sessions collapse into a compact manifold in the \textit{Intent Space}.
Despite divergent packaging, both tactics share the same underlying causal intent \emph{off-platform redirection}, which leads to subsequent actual scams.
By stripping away volatile packaging signals, LPCD isolates this invariant risk core, explaining its robustness to unseen tactical variants.

\subsection{Generality Study~(RQ5)}
\begin{table}[b]
\centering
\caption{Generality study of \textbf{LPCD} across diverse backbone architectures on the \textbf{June OOD} set. Metrics: PR-AUC (AUC), F1-score (F1), R@0.1FPR (R.1), and FPR@0.9R (FPR.9). }
\label{tab:exp:generality}
\resizebox{0.48\textwidth}{!}{
\begin{tabular}{l|c|cccc|c}
\toprule
\textbf{Backbone} & \textbf{Variant} & \small{\textbf{AUC}}$\uparrow$ & \small{\textbf{F1}}$\uparrow$ & \small{\textbf{R.1}}$\uparrow$ & \small{\textbf{FPR.9}}$\downarrow$ & \small{\textbf{Gain (AUC)}} \\
\midrule
\multirow{2}{*}{Transformer} & Vanilla & 0.6208 & 0.5907 & 0.7636 & 0.2545 & -- \\
& \cellcolor{gray!10}+ \textbf{LPCD} & \textbf{0.6573} & \textbf{0.6148} & \textbf{0.7942} & \textbf{0.2232} & \textbf{+5.9\%} \\
\midrule
\multirow{2}{*}{Reformer} & Vanilla & 0.6189 & 0.5967 & 0.7562 & 0.2638 & -- \\
& \cellcolor{gray!10}+ \textbf{LPCD} & \textbf{0.6683} & \textbf{0.6362} & \textbf{0.8172} & \textbf{0.2301} & \textbf{+8.0\%} \\
\midrule
\multirow{2}{*}{TimeMIL} & Vanilla & 0.6316 & 0.5983 & 0.7763 & 0.2288 & -- \\
& \cellcolor{gray!10}+ \textbf{LPCD} &\textbf{0.6779} 
&\textbf{0.6493} 
&\textbf{0.8360} 
&\textbf{0.1949} 
&\textbf{+7.3\%} \\
\midrule
\multirow{2}{*}{TAIL-MIL} & Original & 0.6365 & 0.5869 & 0.7776 & 0.2391 & -- \\
& \cellcolor{gray!10}+ \textbf{LPCD} & \textbf{0.6826} & \textbf{0.6455} & \textbf{0.8327} & \textbf{0.1956} & \textbf{+7.2\%} \\
\bottomrule
\end{tabular}
}
\end{table}
To evaluate the plug-and-play capability of LPCD, we integrate it with diverse backbone architectures, including sequence models (Transformer, Reformer) and MIL-based frameworks (TimeMIL, TAIL-MIL).
As depicted in Table~\ref{tab:exp:generality}, LPCD consistently improves all backbones on the June OOD set. In particular, LPCD achieves +5.9\% to +8.0\% relative PR-AUC gains over the vanilla counterparts, while substantially reducing false positives at high recall.

These consistent gains across both attention-based and pooling-based models indicate that LPCD operates as a model-agnostic plug-in rather than an architecture-dependent design. This suggests that decoupling invariant intent from transient surface behaviors generalizes well across backbone choices and can be applied to existing moderation systems without architectural changes.

\subsection{Online Test~(RQ6)}

\begin{table}[t]
\centering
\caption{Performance on real-world production traffic (01/18/26--01/19/26). Metrics are computed on logs with a 1:10 positive-to-negative sampling ratio. \textbf{LPCD} significantly outperforms the incumbent Transformer and XGBoost models.}
\label{tab:exp:online}
\resizebox{1.0\columnwidth}{!}{
  \begin{tabular}{l|cccc} 
  \toprule
  \textbf{Method} & \textbf{PR-AUC $\uparrow$} & \textbf{F1-score $\uparrow$} & \textbf{R@0.1FPR $\uparrow$} & \textbf{FPR@0.9R $\downarrow$} \\ 
  \midrule
  XGBoost & 0.4229 & 0.4281 & 0.5637 & 0.5779 \\
  Transformer & 0.5855 & 0.6107 & 0.7525 & 0.2287 \\
  \midrule
  \rowcolor{gray!10}
  \textbf{LPCD} & \textbf{0.6578} & \textbf{0.6690} & \textbf{0.8410} & \textbf{0.1625} \\
  \bottomrule
  \end{tabular}
}
\end{table}

To evaluate the real-world impact of LPCD, we evaluate it on the production traffic of a major live streaming platform for A/B testing.
As summarized in Table~\ref{tab:exp:online}, LPCD consistently outperforms the incumbent XGBoost and Transformer models across all metrics, achieving an R@0.1FPR of 0.8410 and a significant reduction in FPR@0.9R (0.1625). These results demonstrate that LPCD's causal decoupling mechanism effectively generalizes to the complex and unpredictable tactical OOD shifts in live environments. By maintaining high precision while suppressing false alarms, LPCD significantly reduces the manual moderation burden and enhances the overall safety of the platform in an industrial-scale deployment.
\section{Conclusion}
In this paper, we identify and address the challenge of \textit{tactical out-of-distribution (OOD) shift} in live streaming risk assessment: a strategic adversarial scenario where malicious actors evolve narrative packaging while maintaining stable objectives. We propose \textbf{LPCD}, a plug-in framework that leverages a latent causal perspective to disentangle invariant intent from volatile packaging. By enforcing latent counterfactual consistency across representative and predictive levels and applying inference-time calibration, LPCD effectively anchors risk detection on stable causal signals, bypassing the need for environment boundaries or raw-level counterfactuals.

Extensive offline experiments and online validation on large-scale industrial traffic demonstrate that LPCD not only achieves superior robustness against evolving tactics but also maintains the efficiency required for real-world moderation. Our work highlights the importance of causal disentanglement in adversarial environments and provides a scalable solution for building robust, intent-focused risk assessment systems.

\begin{acks}
The research work is supported by the National Natural Science Foundation of China under Grant Nos. U2436209, 62576333, and 62406307, the Strategic Priority Research Program of the Chinese Academy of Sciences under Grant No. XDB0680201, the Beijing Natural Science Foundation (F251001), and the Innovation Funding of ICT, CAS under Grant No. E461060.
\end{acks}

\bibliographystyle{ACM-Reference-Format}
\bibliography{sample-sigconf}

@inproceedings{lu2025vlm,
  title={Vlm as policy: Common-law content moderation framework for short video platform},
  author={Lu, Xingyu and Zhang, Tianke and Meng, Chang and Wang, Xiaobei and Wang, Jinpeng and Zhang, Yi-Fan and Tang, Shisong and Liu, Changyi and Ding, Haojie and Jiang, Kaiyu and others},
  booktitle={Proceedings of the 31st ACM SIGKDD Conference on Knowledge Discovery and Data Mining V. 2},
  pages={4682--4693},
  year={2025}
}

@inproceedings{wang2023sequence,
  title={Sequence as genes: an user behavior modeling framework for fraud transaction detection in e-commerce},
  author={Wang, Ziming and Wu, Qianru and Zheng, Baolin and Wang, Junjie and Huang, Kaiyu and Shi, Yanjie},
  booktitle={Proceedings of the 29th ACM SIGKDD Conference on Knowledge Discovery and Data Mining},
  pages={5194--5203},
  year={2023}
}

@inproceedings{xiao2024vecaug,
  title={VecAug: Unveiling Camouflaged Frauds with Cohort Augmentation for Enhanced Detection},
  author={Xiao, Fei and Cai, Shaofeng and Chen, Gang and Jagadish, HV and Ooi, Beng Chin and Zhang, Meihui},
  booktitle={Proceedings of the 30th ACM SIGKDD Conference on Knowledge Discovery and Data Mining},
  pages={6025--6036},
  year={2024}
}

@inproceedings{qiao2025online,
  title={Online Fraud Detection via Test-Time Retrieval-Based Representation Enrichment},
  author={Qiao, Yiran and Wang, Ningtao and Gao, Yuncong and Yang, Yang and Fu, Xing and Wang, Weiqiang and Ao, Xiang},
  booktitle={Proceedings of the AAAI Conference on Artificial Intelligence},
  volume={39},
  number={12},
  pages={12470--12478},
  year={2025}
}

@inproceedings{guo2018learning,
  title={Learning sequential behavior representations for fraud detection},
  author={Guo, Jia and Liu, Guannan and Zuo, Yuan and Wu, Junjie},
  booktitle={2018 IEEE international conference on data mining (ICDM)},
  pages={127--136},
  year={2018},
  organization={IEEE}
}

@inproceedings{wang2025reasoning,
  title={Reasoning-Enhanced Domain-Adaptive Pretraining of Multimodal Large Language Models for Short Video Content Governance},
  author={Wang, Zixuan and Sun, Yu and Wang, Hongwei and Jing, Baoyu and Shen, Xiang and Dong, Xin Luna and Hao, Zhuolin and Xiong, Hongyu and Song, Yang},
  booktitle={Proceedings of the 2025 Conference on Empirical Methods in Natural Language Processing: Industry Track},
  pages={1104--1112},
  year={2025}
}

@inproceedings{lees2022new,
  title={A new generation of perspective api: Efficient multilingual character-level transformers},
  author={Lees, Alyssa and Tran, Vinh Q and Tay, Yi and Sorensen, Jeffrey and Gupta, Jai and Metzler, Donald and Vasserman, Lucy},
  booktitle={Proceedings of the 28th ACM SIGKDD conference on knowledge discovery and data mining},
  pages={3197--3207},
  year={2022}
}

@inproceedings{zannettou2020measuring,
  title={Measuring and characterizing hate speech on news websites},
  author={Zannettou, Savvas and ElSherief, Mai and Belding, Elizabeth and Nilizadeh, Shirin and Stringhini, Gianluca},
  booktitle={Proceedings of the 12th ACM conference on web science},
  pages={125--134},
  year={2020}
}

@INPROCEEDINGS {qiao2024Financial,
author = { Qiao, Yiran and Tang, Yateng and Ao, Xiang and Yuan, Qi and Liu, Ziming and Shen, Chen and Zheng, Xuehao },
booktitle = { 2024 IEEE International Conference on Data Mining (ICDM) },
title = {{ Financial Risk Assessment via Long-term Payment Behavior Sequence Folding }},
year = {2024},
volume = {},
ISSN = {},
pages = {410-419},
doi = {10.1109/ICDM59182.2024.00048},
url = {https://doi.ieeecomputersociety.org/10.1109/ICDM59182.2024.00048},
publisher = {IEEE Computer Society},
address = {Los Alamitos, CA, USA},
month =Dec
}

@inproceedings{dou2020enhancing,
  title={Enhancing graph neural network-based fraud detectors against camouflaged fraudsters},
  author={Dou, Yingtong and Liu, Zhiwei and Sun, Li and Deng, Yutong and Peng, Hao and Yu, Philip S},
  booktitle={Proceedings of the 29th ACM international conference on information \& knowledge management},
  pages={315--324},
  year={2020}
}

@inproceedings{huang2022auc,
  title={Auc-oriented graph neural network for fraud detection},
  author={Huang, Mengda and Liu, Yang and Ao, Xiang and Li, Kuan and Chi, Jianfeng and Feng, Jinghua and Yang, Hao and He, Qing},
  booktitle={Proceedings of the ACM web conference 2022},
  pages={1311--1321},
  year={2022}
}

@inproceedings{shi2022h2,
  title={H2-fdetector: A gnn-based fraud detector with homophilic and heterophilic connections},
  author={Shi, Fengzhao and Cao, Yanan and Shang, Yanmin and Zhou, Yuchen and Zhou, Chuan and Wu, Jia},
  booktitle={Proceedings of the ACM web conference 2022},
  pages={1486--1494},
  year={2022}
}

@inproceedings{li2021live,
  title={Live-streaming fraud detection: A heterogeneous graph neural network approach},
  author={Li, Zhao and Wang, Haishuai and Zhang, Peng and Hui, Pengrui and Huang, Jiaming and Liao, Jian and Zhang, Ji and Bu, Jiajun},
  booktitle={Proceedings of the 27th ACM SIGKDD Conference on Knowledge Discovery \& Data Mining},
  pages={3670--3678},
  year={2021}
}

@article{cheng2025graph,
  title={Graph neural networks for financial fraud detection: a review},
  author={Cheng, Dawei and Zou, Yao and Xiang, Sheng and Jiang, Changjun},
  journal={Frontiers of Computer Science},
  volume={19},
  number={9},
  pages={1--15},
  year={2025},
  publisher={Springer}
}

@article{arjovsky2019invariant,
  title={Invariant risk minimization},
  author={Arjovsky, Martin and Bottou, L{\'e}on and Gulrajani, Ishaan and Lopez-Paz, David},
  journal={arXiv preprint arXiv:1907.02893},
  year={2019}
}

@inproceedings{krueger2021out,
  title={Out-of-distribution generalization via risk extrapolation (rex)},
  author={Krueger, David and Caballero, Ethan and Jacobsen, Joern-Henrik and Zhang, Amy and Binas, Jonathan and Zhang, Dinghuai and Le Priol, Remi and Courville, Aaron},
  booktitle={International conference on machine learning},
  pages={5815--5826},
  year={2021},
  organization={PMLR}
}

@article{zhou2022domain,
  title={Domain generalization: A survey},
  author={Zhou, Kaiyang and Liu, Ziwei and Qiao, Yu and Xiang, Tao and Loy, Chen Change},
  journal={IEEE transactions on pattern analysis and machine intelligence},
  volume={45},
  number={4},
  pages={4396--4415},
  year={2022},
  publisher={IEEE}
}

@article{liu2021towards,
  title={Towards out-of-distribution generalization: A survey},
  author={Liu, Jiashuo and Shen, Zheyan and He, Yue and Zhang, Xingxuan and Xu, Renzhe and Yu, Han and Cui, Peng},
  journal={arXiv preprint arXiv:2108.13624},
  year={2021}
}

@article{zhang2020causal,
  title={A causal view on robustness of neural networks},
  author={Zhang, Cheng and Zhang, Kun and Li, Yingzhen},
  journal={Advances in Neural Information Processing Systems},
  volume={33},
  pages={289--301},
  year={2020}
}

@article{liu2021learning,
  title={Learning causal semantic representation for out-of-distribution prediction},
  author={Liu, Chang and Sun, Xinwei and Wang, Jindong and Tang, Haoyue and Li, Tao and Qin, Tao and Chen, Wei and Liu, Tie-Yan},
  journal={Advances in Neural Information Processing Systems},
  volume={34},
  pages={6155--6170},
  year={2021}
}

@inproceedings{mahajan2021domain,
  title={Domain generalization using causal matching},
  author={Mahajan, Divyat and Tople, Shruti and Sharma, Amit},
  booktitle={International conference on machine learning},
  pages={7313--7324},
  year={2021},
  organization={PMLR}
}

@inproceedings{oublal2024disentangling,
  title={Disentangling time series representations via contrastive independence-of-support on l-variational inference},
  author={Oublal, Khalid and Ladjal, Said and Benhaiem, David and LE BORGNE, Emmanuel and Roueff, Fran{\c{c}}ois},
  booktitle={The Twelfth International Conference on Learning Representations},
  year={2024}
}

@article{liu2025long,
  title={Long-term urban flow prediction against data distribution shift: A causal perspective},
  author={Liu, Yuting and Zhou, Qiang and Li, Hanzhe and Zhuang, Fuzhen and Gu, Jingjing},
  journal={IEEE Transactions on Knowledge and Data Engineering},
  year={2025},
  publisher={IEEE}
}

@article{wu2025out,
  title={Out-of-distribution generalization in time series: A survey},
  author={Wu, Xin and Teng, Fei and Li, Xingwang and Zhang, Ji and Duan, Qiang and Li, Tianrui},
  journal={Information Fusion},
  pages={104336},
  year={2026},
  publisher={Elsevier}
}

@article{feder2022causal,
  title={Causal inference in natural language processing: Estimation, prediction, interpretation and beyond},
  author={Feder, Amir and Keith, Katherine A and Manzoor, Emaad and Pryzant, Reid and Sridhar, Dhanya and Wood-Doughty, Zach and Eisenstein, Jacob and Grimmer, Justin and Reichart, Roi and Roberts, Margaret E and others},
  journal={Transactions of the Association for Computational Linguistics},
  volume={10},
  pages={1138--1158},
  year={2022},
  publisher={MIT Press One Broadway, 12th Floor, Cambridge, Massachusetts 02142, USA~…}
}

@inproceedings{higgins2017beta,
  title={beta-vae: Learning basic visual concepts with a constrained variational framework},
  author={Higgins, Irina and Matthey, Loic and Pal, Arka and Burgess, Christopher and Glorot, Xavier and Botvinick, Matthew and Mohamed, Shakir and Lerchner, Alexander},
  booktitle={International conference on learning representations},
  year={2017}
}

@inproceedings{kim2018disentangling,
  title={Disentangling by factorising},
  author={Kim, Hyunjik and Mnih, Andriy},
  booktitle={International conference on machine learning},
  pages={2649--2658},
  year={2018},
  organization={PMLR}
}

@article{bousmalis2016domain,
  title={Domain separation networks},
  author={Bousmalis, Konstantinos and Trigeorgis, George and Silberman, Nathan and Krishnan, Dilip and Erhan, Dumitru},
  journal={Advances in neural information processing systems},
  volume={29},
  year={2016}
}

@book{pearl2009causality,
  title={Causality},
  author={Pearl, Judea},
  year={2009},
  publisher={Cambridge university press}
}

@inproceedings{schroff2015facenet,
  title={Facenet: A unified embedding for face recognition and clustering},
  author={Schroff, Florian and Kalenichenko, Dmitry and Philbin, James},
  booktitle={Proceedings of the IEEE conference on computer vision and pattern recognition},
  pages={815--823},
  year={2015}
}

@inproceedings{sauercounterfactual,
  title={Counterfactual Generative Networks},
  author={Sauer, Axel and Geiger, Andreas},
  booktitle={International Conference on Learning Representations},
  year = {2021}
}

@article{li2018adaptive,
  title={Adaptive batch normalization for practical domain adaptation},
  author={Li, Yanghao and Wang, Naiyan and Shi, Jianping and Hou, Xiaodi and Liu, Jiaying},
  journal={Pattern Recognition},
  volume={80},
  pages={109--117},
  year={2018},
  publisher={Elsevier}
}

@inproceedings{chen2020simclr,
  title={Simclr: A simple framework for contrastive learning of visual representations},
  author={Chen, Ting and Kornblith, Simon and Norouzi, Mohammad and Hinton, Geoffrey},
  booktitle={International Conference on Learning Representations},
  volume={2},
  number={4},
  year={2020},
  organization={PMLR New York, NY, USA}
}

@article{veitch2021counterfactual,
  title={Counterfactual invariance to spurious correlations in text classification},
  author={Veitch, Victor and D'Amour, Alexander and Yadlowsky, Steve and Eisenstein, Jacob},
  journal={Advances in neural information processing systems},
  volume={34},
  pages={16196--16208},
  year={2021}
}

@article{vaswani2017attention,
  title={Attention is all you need},
  author={Vaswani, Ashish and Shazeer, Noam and Parmar, Niki and Uszkoreit, Jakob and Jones, Llion and Gomez, Aidan N and Kaiser, {\L}ukasz and Polosukhin, Illia},
  journal={Advances in neural information processing systems},
  volume={30},
  year={2017}
}

@article{kitaev2020reformer,
  title={Reformer: The efficient transformer},
  author={Kitaev, Nikita and Kaiser, {\L}ukasz and Levskaya, Anselm},
  journal={arXiv preprint arXiv:2001.04451},
  year={2020}
}

@inproceedings{zhou2021informer,
  title={Informer: Beyond efficient transformer for long sequence time-series forecasting},
  author={Zhou, Haoyi and Zhang, Shanghang and Peng, Jieqi and Zhang, Shuai and Li, Jianxin and Xiong, Hui and Zhang, Wancai},
  booktitle={Proceedings of the AAAI conference on artificial intelligence},
  volume={35},
  number={12},
  pages={11106--11115},
  year={2021}
}

@inproceedings{early2024inherently,
  title={Inherently Interpretable Time Series Classification via Multiple Instance Learning},
  author={Early, Joseph and Cheung, Gavin KC and Cutajar, Kurt and Xie, Hanting and Kandola, Jas and Twomey, Niall},
  booktitle={ICLR},
  year={2024}
}

@inproceedings{chen2024timemil,
  title={TimeMIL: advancing multivariate time series classification via a time-aware multiple instance learning},
  author={Chen, Xiwen and Qiu, Peijie and Zhu, Wenhui and Li, Huayu and Wang, Hao and Sotiras, Aristeidis and Wang, Yalin and Razi, Abolfazl},
  booktitle={Proceedings of the 41st International Conference on Machine Learning},
  pages={7190--7206},
  year={2024}
}

@inproceedings{jang2025tail,
  title={TAIL-MIL: Time-aware and instance-learnable multiple instance learning for multivariate time series anomaly detection},
  author={Jang, Jaeseok and Kwon, Hyuk-Yoon},
  booktitle={Proceedings of the AAAI Conference on Artificial Intelligence},
  volume={39},
  number={17},
  pages={17582--17589},
  year={2025}
}

@article{ahuja2021invariance,
  title={Invariance principle meets information bottleneck for out-of-distribution generalization},
  author={Ahuja, Kartik and Caballero, Ethan and Zhang, Dinghuai and Gagnon-Audet, Jean-Christophe and Bengio, Yoshua and Mitliagkas, Ioannis and Rish, Irina},
  journal={Advances in Neural Information Processing Systems},
  volume={34},
  pages={3438--3450},
  year={2021}
}

@article{yan2020improve,
  title={Improve unsupervised domain adaptation with mixup training},
  author={Yan, Shen and Song, Huan and Li, Nanxiang and Zou, Lincan and Ren, Liu},
  journal={arXiv preprint arXiv:2001.00677},
  year={2020}
}

@inproceedings{sun2016deep,
  title={Deep coral: Correlation alignment for deep domain adaptation},
  author={Sun, Baochen and Saenko, Kate},
  booktitle={European conference on computer vision},
  pages={443--450},
  year={2016},
  organization={Springer}
}

@inproceedings{
Sagawa*2020Distributionally,
title={Distributionally Robust Neural Networks},
author={Shiori Sagawa* and Pang Wei Koh* and Tatsunori B. Hashimoto and Percy Liang},
booktitle={International Conference on Learning Representations},
year={2020},
url={https://openreview.net/forum?id=ryxGuJrFvS}
}

@inproceedings{kim2025sufficient,
  title={Sufficient invariant learning for distribution shift},
  author={Kim, Taero and Park, Subeen and Lim, Sungjun and Jung, Yonghan and Muandet, Krikamol and Song, Kyungwoo},
  booktitle={Proceedings of the Computer Vision and Pattern Recognition Conference},
  pages={4958--4967},
  year={2025}
}

@inproceedings{creager2021environment,
  title={Environment inference for invariant learning},
  author={Creager, Elliot and Jacobsen, J{\"o}rn-Henrik and Zemel, Richard},
  booktitle={International Conference on Machine Learning},
  pages={2189--2200},
  year={2021},
  organization={PMLR}
}

@inproceedings{liu2024time,
  title={Time-series forecasting for out-of-distribution generalization using invariant learning},
  author={Liu, Haoxin and Kamarthi, Harshavardhan and Kong, Lingkai and Zhao, Zhiyuan and Zhang, Chao and Prakash, B Aditya},
  booktitle={Proceedings of the 41st International Conference on Machine Learning},
  pages={31312--31325},
  year={2024}
}

@inproceedings{
loshchilov2018decoupled,
title={Decoupled Weight Decay Regularization},
author={Ilya Loshchilov and Frank Hutter},
booktitle={International Conference on Learning Representations},
year={2019},
url={https://openreview.net/forum?id=Bkg6RiCqY7},
}

@inproceedings{qiao2026livelieactionawarecapsule,
  title={Live or Lie: Action-Aware Capsule Multiple Instance Learning for Risk Assessment in Live Streaming Platforms},
  author={Qiao, Yiran and Chen, Jing and Ao, Xiang and Zhong, Qiwei and Liu, Yang and He, Qing},
  booktitle={Proceedings of the 32nd ACM SIGKDD Conference on Knowledge Discovery and Data Mining V. 1},
  pages={1182--1193},
  year={2026}
}

\appendix
\section{Baseline Details}
\label{sec:app:exp}

First, we adopt two categories of backbone models as candidates to validate the effectiveness of LPCD.
(i)~\textit{Sequence Models} explicitly model the action sequences of sessions: 

\begin{itemize}[leftmargin=*,topsep=5pt]

\item

\textbf{Transformer}~\cite{vaswani2017attention} serves as a standard self-attention baseline. 

\item 

\textbf{Reformer}~\cite{kitaev2020reformer} improves efficiency via locality-sensitive hashing. 

\item 

\textbf{Informer}~\cite{zhou2021informer} further scales to long sequences through sparse attention and representation distillation.

\end{itemize}

\noindent
(ii) \textit{MIL methods} aggregate instance-level signals into session-level predictions, where each instance corresponds to a per-user action subsequence within a 100-second window: 

\begin{itemize}[leftmargin=*,topsep=5pt]

\item \textbf{MIL-LET}~\cite{early2024inherently} introduces an MIL formulation for time-series classification that provides localized interpretability.
\item \textbf{TimeMIL}~\cite{chen2024timemil} introduces temporal awareness via learnable wavelet-based positional encodings. 
\item \textbf{TAIL-MIL}~\cite{jang2025tail} extends MIL to multivariate time-series modeling using a 2D formulation.
\item \textbf{AC-MIL}~\cite{qiao2026livelieactionawarecapsule} is a domain-specific MIL framework for live-streaming risk assessment that jointly models user-level and temporal patterns.

\end{itemize}

Second, we compare LPCD with four types of plug-in methods for OOD generalization to show its superiority. 
(i) \textit{Invariant Learning (IL)} methods aim to capture invariant causal relationships across different environments by penalizing unstable correlations:
\begin{itemize}[leftmargin=*,topsep=5pt]
\item \textbf{IRM}~\cite{arjovsky2019invariant} introduces a gradient-based penalty to ensure the optimal classifier is consistent across all training environments.
\item \textbf{VREx}~\cite{krueger2021out} reduces the variance of risks across environments to achieve better generalization under distribution shifts.
\item \textbf{IB-IRM}~\cite{ahuja2021invariance} combines the Information Bottleneck principle with IRM to filter out environment-specific noise while preserving invariant features.
\end{itemize}

\noindent
(ii) \textit{Data Augmentation and Alignment (DA)} methods focus on enhancing model robustness by expanding the training distribution or aligning feature-level statistics:
\begin{itemize}[leftmargin=*,topsep=5pt]
    \item \textbf{Mixup}~\cite{yan2020improve} creates vicinal training samples through linear interpolation of feature-label pairs to smooth decision boundaries.
    \item \textbf{CORAL}~\cite{sun2016deep} aligns the second-order statistics (covariance) of source and target domain distributions to learn domain-invariant representations.
\end{itemize}

\noindent
(iii) \textit{Distributionally Robust Optimization (DRO)} methods optimize for the worst-case performance across groups to mitigate spurious correlations and enhance stability:
\begin{itemize}[leftmargin=*,topsep=5pt]
\item \textbf{GroupDRO}~\cite{Sagawa*2020Distributionally} explicitly minimizes the maximum loss across different groups to mitigate the impact of spurious correlations.
\item \textbf{ASGDRO}~\cite{kim2025sufficient} seeks common flat minima across environments to learn a diverse set of invariant features.
\end{itemize}

\noindent
(iv) \textit{Environment Inference (EI)} methods tackle the challenge of missing environment labels by automatically discovering latent environmental structures:
\begin{itemize}[leftmargin=*,topsep=5pt]
\item \textbf{EIIL}~\cite{creager2021environment} infers environments by searching for a partition that maximally violates the IRM invariance principle.
\item \textbf{FOIL}~\cite{liu2024time} identifies latent environments in time-series data by optimizing for feature-level stability over temporal segments.
\end{itemize}

It is worth noting that methods in the first three categories (IL, DA, and DRO) rely on explicit environment annotations during training, whereas EI methods and our proposed LPCD operate without any prior environmental labels.

Since no ground-truth environment annotations are available, we follow common practice~\cite{creager2021environment} and construct training environments via temporal partitioning. Specifically, for the \textbf{May} dataset, the training period (05/20--06/03) is divided into four distinct environments: \textit{May 20--23}, \textit{May 24--27}, \textit{May 28--31}, and \textit{June 1--3}. For the \textbf{June} dataset, the training window (06/04--06/10) is partitioned into three environments: \textit{June 4--6}, \textit{June 7--8}, and \textit{June 9--10}.

\section{Supplementary Experiments}
\subsection{Hyperparameter Sensitivity Test}
\label{sec:app:hyper}
\begin{figure}[htbp]
    \centering
    \includegraphics[width=\linewidth]{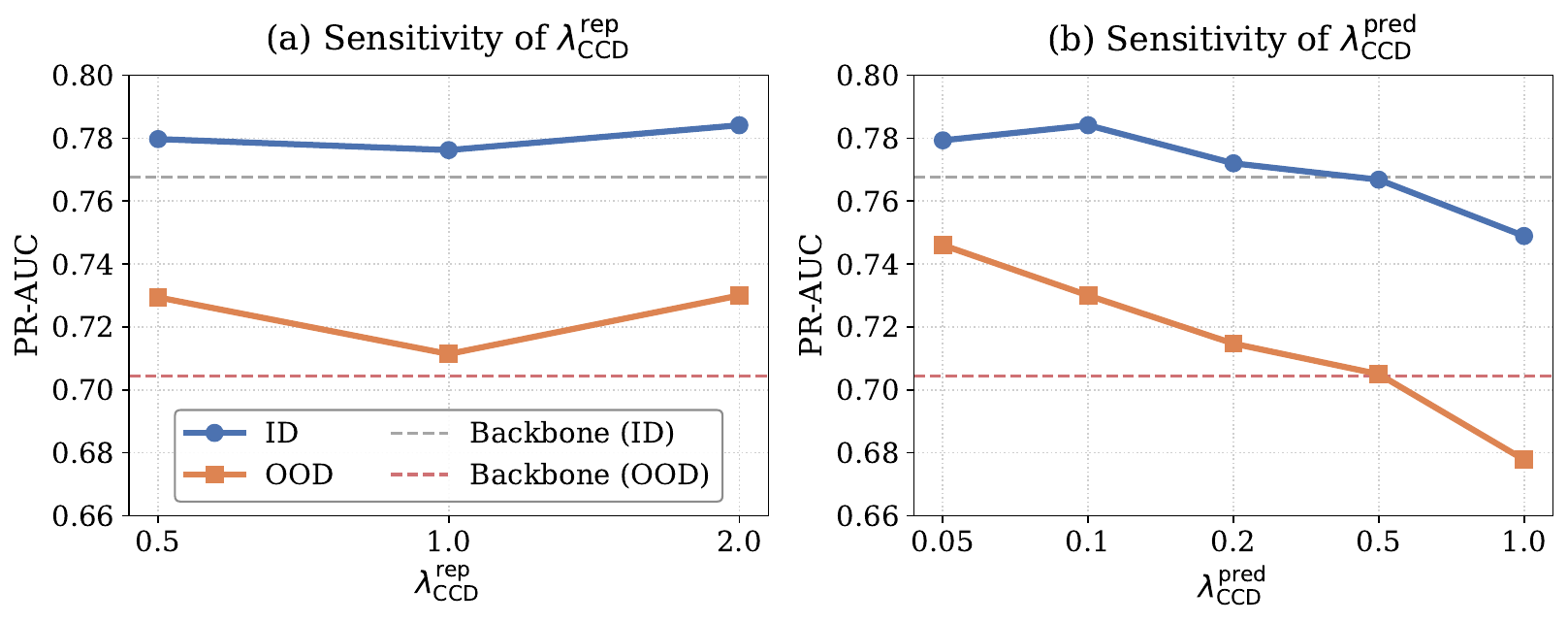} 
    \caption{Hyperparameter sensitivity analysis on the May dataset. Subplots (a) and (b) illustrate the impact of $\lambda_{\mathrm{CCD}}^{\mathrm{rep}}$ and $\lambda_{\mathrm{CCD}}^{\mathrm{pred}}$ on PR-AUC, respectively. Solid lines with markers represent LPCD performance, while dashed horizontal lines represent the corresponding AC-MIL backbone baselines for ID (blue) and OOD (red) test sets.}
    \label{fig:app:hyper}
\end{figure}

We evaluate the sensitivity of our LPCD framework to the two balance hyperparameters in the CCD module: $\lambda_{\mathrm{CCD}}^{\mathrm{rep}}$ and $\lambda_{\mathrm{CCD}}^{\mathrm{pred}}$. The experiments are conducted on the May dataset, and the results are summarized in Figure~\ref{fig:app:hyper}.

\textbf{Sensitivity of $\lambda_{\mathrm{CCD}}^{\mathrm{rep}}$:} As shown in Figure~\ref{fig:app:hyper}(a), the model performance in both In-ID and Tactical OOD scenarios remains consistently higher than the AC-MIL backbone across all tested values. The PR-AUC for OOD reaches its peak at $\lambda_{\mathrm{CCD}}^{\mathrm{rep}}=2.0$ (0.7300), demonstrating that representation-level consistency is robust to varying regularization strengths.
\textbf{Sensitivity of $\lambda_{\mathrm{CCD}}^{\mathrm{pred}}$:} Figure~\ref{fig:app:hyper}(b) reveals that the model is more sensitive to the prediction-level consistency weight. While small values yield the best OOD performance (peaking at 0.7460 with $\lambda_{\mathrm{CCD}}^{\mathrm{pred}}=0.05$), an excessive penalty (e.g., 1.0) leads to performance decay. This suggests that while predictive consistency helps in decoupling, an excessive penalty may overly constrain the decision boundary.

\subsection{Analysis of Post-hoc Calibration Variants}
\label{sec:app:ab}
To evaluate the effectiveness of our proposed \textbf{Dimensional Magnitude Alignment (V0)}, which serves as the Post-hoc Magnitude Calibration module (Section~\ref{sec:met:cal}), we compare it with four alternative parameter-free calibration variants. These variants are designed to rectify distributional shifts in the packaging manifold $\mathbf{z}_{\mathrm{pack}}$ as follows:

\begin{itemize}[leftmargin=*,topsep=5pt]
    \item \textbf{V0 (Dimensional Magnitude Alignment - Default):} Performs per-dimension rescaling using a diagonal matrix $\boldsymbol{\Gamma}$: $\tilde{\mathbf{z}}_{\mathrm{pack}} = \boldsymbol{\Gamma} \mathbf{z}_{\mathrm{pack}}$, where $\gamma_d = \sigma_{\mathrm{train}, d} / \sigma_{\mathrm{test}, d}$. It targets anisotropic magnitude shifts in specific latent dimensions.
    
    \item \textbf{V1 (Instance Norm Rescaling):} A sample-level constraint that forces the $L_2$ norm of each representation to match the training average $r_{\mathrm{train}}$: $\tilde{\mathbf{z}}_{\mathrm{pack}} = \mathbf{z}_{\mathrm{pack}} \cdot (r_{\mathrm{train}} / \|\mathbf{z}_{\mathrm{pack}}\|_2)$. It ensures energy consistency but ignores dimensional variance.
    
    \item \textbf{V2 (Counterfactual Consistency Check):} A reasoning-level check that compares the factual prediction with a counterfactual one wrapping the same intent in a pre-defined safe prototype $\bar{\mathbf{z}}_{\mathrm{pack}}^{\mathrm{safe}}$. The final risk probability is: $\hat{y}_{\mathrm{final}} = \min( \hat{y}_{\mathrm{fact}}, \hat{y}_{\mathrm{cf}} )$.
    
    \item \textbf{V3 (Centroid Translation Alignment):} A distribution-level translation that eliminates systemic bias by subtracting the mean drift: $\tilde{\mathbf{z}}_{\mathrm{pack}} = \mathbf{z}_{\mathrm{pack}} - (\mu_{\mathrm{test}} - \mu_{\mathrm{train}})$, where $\mu$ denotes the centroid of the packaging manifold.
    
    \item \textbf{V4 (Second-order Correlation Alignment):} A rigorous affine transformation that synchronizes both mean and covariance ($\Sigma$): $\tilde{\mathbf{z}}_{\mathrm{pack}} = \Sigma_{\mathrm{train}}^{1/2} \Sigma_{\mathrm{test}}^{-1/2} (\mathbf{z}_{\mathrm{pack}} - \mu_{\mathrm{test}}) + \mu_{\mathrm{train}}$.
\end{itemize}

\begin{table}[b]
\centering
\caption{Comparison of parameter-free calibration variants on the June OOD Test Set. All variants are applied to the frozen LPCD architecture. \textbf{V0} is the default strategy. Metrics: PR-AUC (AUC), F1-score (F1), R@0.1FPR (R.1), and FPR@0.9R (FPR.9).}
\label{tab:appendix:calib}
\resizebox{0.48\textwidth}{!}{ 
\begin{tabular}{l|c|cccc}
\toprule
\textbf{Calibration Variant} & \textbf{Level} & \textbf{AUC}$\uparrow$ & \textbf{F1}$\uparrow$ & \textbf{R.1}$\uparrow$ & \textbf{FPR.9}$\downarrow$ \\
\midrule
No Calibration (LPCD) & - & 0.7053 & 0.6388 & 0.8178 & 0.2041 \\
\midrule
V1 (Instance Norm Rescaling) & Sample & 0.7061 & 0.6008 & 0.8192 & 0.1994 \\
V2 (Counterfactual Consistency) & Reasoning & 0.7055 & 0.6388 & 0.8178 & 0.2041 \\
V3 (Centroid Translation) & Distribution & 0.7053 & 0.6362 & 0.8178 & 0.2040 \\
V4 (Second-order Correlation) & Distribution & 0.7051 & 0.6361 & 0.8185 & 0.2030 \\
\midrule
\rowcolor{gray!10} \textbf{V0 (Dimensional Magnitude)} & \textbf{Dimension} & \textbf{0.7287} & \textbf{0.6779} & \textbf{0.8600} & \textbf{0.1732} \\
\bottomrule
\end{tabular}
}
\end{table}

\noindent
\textbf{Analysis of Results.} 
As shown in Table~\ref{tab:appendix:calib}, \textbf{V0} significantly outperforms all other variants, from which we derive two key insights:
(1) \textbf{Dimension-specific sensitivity:} The performance degradation of V1 in F1-score (0.6008 vs. 0.6388) suggests that global scalar scaling destroys the relative importance across different latent dimensions. In our disentangled space, dimensions carry independent semantic signals; forcing a uniform norm introduces excessive noise and distorts the discriminative structure.
(2) \textbf{Effective factor decorrelation:} The marginal gains of V4 over V3 indicate that the orthogonality constraint ($\mathcal{L}_{\mathrm{ortho}}$) during training successfully minimized cross-dimensional correlations. Consequently, complex covariance-based alignment collapses toward simpler mean alignment. This underscores that \emph{magnitude shift}, rather than rotational or correlation shift, is the primary bottleneck in OOD deployment, which V0 addresses with optimal granularity.
\balance
\end{document}